\documentclass[journal]{IEEEtran}
\usepackage{algorithm}
\usepackage{algorithmicx}
\usepackage{algpseudocode}
\algnewcommand{\LineComment}[1]{\State \(\triangleright\) #1}
\usepackage{amssymb}
\usepackage{bbm}
\usepackage[mathscr]{euscript}
\usepackage{multirow}
\usepackage{booktabs}
\usepackage[dvipsnames]{xcolor}
\usepackage{url}
\usepackage{makecell}

\usepackage[backref]{hyperref}
\hypersetup{
    colorlinks=true,
    linkcolor=blue,
    filecolor=magenta,      
    urlcolor=cyan
}

\usepackage{epsfig}
\usepackage{graphicx}
\usepackage{subdefs}

\usepackage{colortbl}
\definecolor{Gray}{gray}{0.85}
\definecolor{SkyBlue}{rgb}{0.88,1,1}
\definecolor{PasteGreen}{RGB}{204, 226, 215}
\definecolor{PastePink}{RGB}{253, 223, 236}
\definecolor{PasteYellow}{RGB}{254, 255, 214}
\definecolor{PasteLavender}{RGB}{229, 228, 244}
\definecolor{NewGreen}{RGB}{17,207,156}
\newcolumntype{a}{>{\columncolor{Gray}}c}

\ifCLASSINFOpdf
\else
\fi

\usepackage{amsmath}
\begin{document}
%
\title{Counting in the 2020s: Binned Representations and Inclusive Performance Measures for Deep Crowd Counting Approaches}
%
%
%


\author{Sravya~Vardhani~Shivapuja,
        Ashwin~Gopinath,
        Ayush Gupta,
        Ganesh Ramakrishnan,
        and~Ravi~Kiran~Sarvadevabhatla,~\IEEEmembership{Member,~IEEE}

\thanks{M. Shell was with the Department
of Electrical and Computer Engineering, Georgia Institute of Technology, Atlanta,
GA, 30332 USA e-mail: (see http://www.michaelshell.org/contact.html).}
\thanks{J. Doe and J. Doe are with Anonymous University.}
\thanks{Manuscript received April 19, 2005; revised August 26, 2015.}}

%
%

\markboth{Journal of \LaTeX\ Class Files,~Vol.~14, No.~8, August~2015}%
{Shell \MakeLowercase{\textit{et al.}}: The right way in Crowd Counting?}
%



\maketitle


\begin{abstract}
The data distribution in popular crowd counting datasets is typically heavy tailed and discontinuous. This skew affects all stages within the pipelines of deep crowd counting approaches. Specifically, the approaches exhibit unacceptably large standard deviation wrt statistical measures (MSE, MAE). To address such concerns in a holistic manner, we make two fundamental contributions. Firstly, we modify the training pipeline to accommodate the knowledge of dataset skew. To enable principled and balanced minibatch sampling, we propose a novel smoothed Bayesian binning approach. More specifically, we propose a novel cost function which can be readily incorporated into existing crowd counting deep networks to encourage bin-aware optimization. 
As the second contribution, we introduce additional performance measures which are more inclusive and throw light on various comparative performance aspects of the deep networks. We also show that our binning-based modifications retain their superiority wrt the newly proposed performance measures. Overall, our contributions enable a practically useful and detail-oriented characterization of performance for crowd counting approaches.
    

\end{abstract}

%
\IEEEpeerreviewmaketitle

\section{Introduction}
\label{section:intro}

Given an image, deep counting networks regress a single value representing the number of people in the image. Estimating count people from images has significant applications in urban planning, surveillance in industries, hospitals and other establishments. Deep networks in crowd counting are typically trained on images and ground truth annotations which could be image pixel coordinates (or point annotations) of heads or density maps prepared using these coordinates. The output for these networks is predicted count estimates, either through predicted coordinates of heads or through predicted density maps.

\begin{figure}[!t]
\centering
\includegraphics[width=0.5\textwidth]{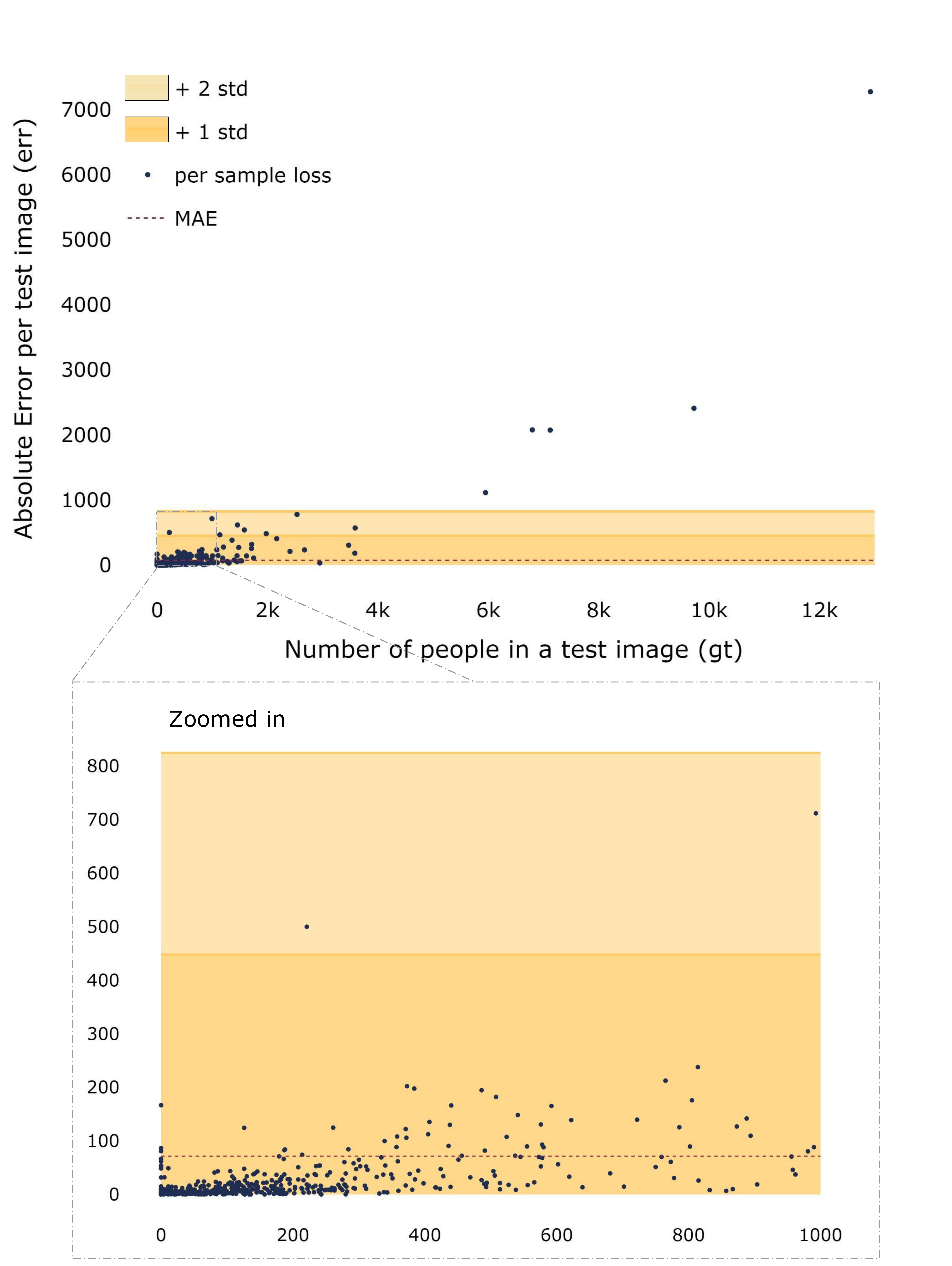}
\caption{The scatter plot of ground-truth counts and absolute errors by DM-Count~\cite{wang2020DMCount} on the NWPU dataset~\cite{gao2020nwpu}. The Mean Absolute Error (MAE) is $71.71$, but the standard deviation is multiple orders of magnitude larger: $376.40$. The zoomed in plot shows that even for lowest count (0 people), error is significantly larger than $0$ (~150). Clearly, MAE is a poor representative of performance across count range.}
\label{fig:1}
\end{figure}

Recent large-scale datasets used to train deep counting networks include Shanghai Tech~\cite{mcnn}, UCF-QNRF~\cite{idrees2018composition}, JHU~\cite{9009496} and NWPU-Crowd~\cite{gao2020nwpu}. These datasets have considerably helped advance the state-of-the-art in crowd counting approaches. However, the data distribution in these datasets tends to be heavy-tailed. In other words, they contain a large number of images with small (people) count and a rather limited number of images with a large count (see Figure~\ref{fig:1}). In addition, discontinuities ({\em i.e.}, lack of samples) can also be seen in the data distribution. 

\begin{table*}[!t]
\caption{Summary of the Networks. The grouping of the networks is based on their ground truth and output. Different groups are assigned with different colors.}
\label{table:summarymodels}
\centering
{
\centering
\begin{tabular}{c|p{2cm}|p{8cm}|p{2cm}|p{2cm}}
\toprule
Model Name & Conference, & Model Summary & Ground truth & Output \\
 &  Year &  & & \\
\toprule

\cellcolor{PasteLavender}&\cellcolor{PasteLavender} &\cellcolor{PasteLavender} &\cellcolor{PasteLavender} &\cellcolor{PasteLavender} \\

\cellcolor{PasteLavender} P2PNet~\cite{Song_2021_ICCV} &\cellcolor{PasteLavender} ICCV, \par 2021

& \cellcolor{PasteLavender} A first-of-a-kind Purely point based framework for crowd counting and localization. This solution predicts a set of point locations consistent with the ground truth point annotations. In order to accomplish the task, the novel idea is to assign learning targets for point proposals which has been implemented using Hungarian algorithm. 

&  \cellcolor{PasteLavender} Point annotations & \cellcolor{PasteLavender} Count estimated from Point Annotations \\
\cellcolor{PasteLavender}&\cellcolor{PasteLavender} &\cellcolor{PasteLavender} &\cellcolor{PasteLavender} &\cellcolor{PasteLavender} \\
 
\cellcolor{PasteYellow} &\cellcolor{PasteYellow} &\cellcolor{PasteYellow} &\cellcolor{PasteYellow} &\cellcolor{PasteYellow} \\

\cellcolor{PasteYellow} FIDTM~\cite{liang2021focal} &\cellcolor{PasteYellow} arXiv,  \par 2021
& \cellcolor{PasteYellow} To avoid location imprecision and inaccuracy caused by dense overlaps due to Gaussian blurs, this model uses a novel map based on Focal Inverse Distance Transform. & \cellcolor{PasteYellow} Point annotations & \cellcolor{PasteYellow} Count estimated from \par Focal Inverse Distance Map \\

\cellcolor{PasteYellow} &\cellcolor{PasteYellow} &\cellcolor{PasteYellow} &\cellcolor{PasteYellow} &\cellcolor{PasteYellow} \\

\cellcolor{PasteGreen} &\cellcolor{PasteGreen} &\cellcolor{PasteGreen} &\cellcolor{PasteGreen} &\cellcolor{PasteGreen} \\

\cellcolor{PasteGreen} GenLoss~\cite{Wan_2021_CVPR}&\cellcolor{PasteGreen} CVPR, \par 2021 
& \cellcolor{PasteGreen} Learning density map representation is modelled as an unbalanced optimal transport problem. A generalized loss is proposed has pixel-wise L2 loss and Bayesian loss~\cite{ma2019bayesian} as sub-optimal special cases. A perspective-guided cost function is used to deal with the perspective information.  & \cellcolor{PasteGreen} Point Annotations & \cellcolor{PasteGreen} Count estimated from \par Density Maps \\
\cellcolor{PasteGreen} &\cellcolor{PasteGreen} &\cellcolor{PasteGreen} &\cellcolor{PasteGreen} &\cellcolor{PasteGreen} \\

\cellcolor{PasteGreen}DM-Count~\cite{wang2020DMCount} &\cellcolor{PasteGreen}NeurlPS, \par 2020 
& \cellcolor{PasteGreen} Uses Distribution Matching for crowd COUNTing (DM-Count). An Optimal Transport (OT) Loss is used to find the similarity between normalized predicted density map and the normalized ground truth density map. To stabilize this OT loss, a Total Variation loss is used. & \cellcolor{PasteGreen} Point Annotations & \cellcolor{PasteGreen} Count estimated from \par Density Maps \\
\cellcolor{PasteGreen} &\cellcolor{PasteGreen} &\cellcolor{PasteGreen} &\cellcolor{PasteGreen} &\cellcolor{PasteGreen} \\

\cellcolor{PasteGreen}BL~\cite{ma2019bayesian} &\cellcolor{PasteGreen}ICCV, \par  2019 
& \cellcolor{PasteGreen} To avoid an imperfect “ground-truth” density map which has occlusions,  perspective effects, variations in object shapes, {\em etc.}, this model uses a novel loss function which constructs a density contribution probability model from the point annotations at every pixel.  & \cellcolor{PasteGreen} Point Annotations & \cellcolor{PasteGreen} Count estimated from \par Density Maps \\
\cellcolor{PasteGreen} &\cellcolor{PasteGreen} &\cellcolor{PasteGreen} &\cellcolor{PasteGreen} &\cellcolor{PasteGreen} \\

\cellcolor{PastePink} &\cellcolor{PastePink} &\cellcolor{PastePink} &\cellcolor{PastePink} &\cellcolor{PastePink} \\

\cellcolor{PastePink}SASNet~\cite{sasnet} &\cellcolor{PastePink}AAAI, \par  2021 
& \cellcolor{PastePink} To avoid large scale feature variation problem in crowd counting settings, multi-scale feature representations have been used in multi-level network. Novel PRA loss is introduced to further alleviate the over-estimated or under-estimated region. & \cellcolor{PastePink} Density maps constructed from Point annotations & \cellcolor{PastePink} Count estimated from \par Density Maps \\
\cellcolor{PastePink} &\cellcolor{PastePink} &\cellcolor{PastePink} &\cellcolor{PastePink} &\cellcolor{PastePink} \\

\cellcolor{PastePink}S-DCNet~\cite{xhp2019SDCNet} &\cellcolor{PastePink}ICCV, \par  2019 
& \cellcolor{PastePink} Crowd counting task is compared to an open-set problem {\em i.e} the count in the image can vary from $[0,+\infty)$. During training however, there is only a smaller closed set observed by the network. Their idea is to build a network that trains on closed set and generalizes well on the open set (unobserved).


& \cellcolor{PastePink} Density maps constructed from Point annotations & \cellcolor{PastePink} Count estimated from \par Density Maps \\
\cellcolor{PastePink} &\cellcolor{PastePink} &\cellcolor{PastePink} &\cellcolor{PastePink} &\cellcolor{PastePink} \\

\cellcolor{PastePink}SCAR~\cite{gao2019scar} &\cellcolor{PastePink} Neurocomputing \par Journal,  \par 2019 
& \cellcolor{PastePink} In this work, a Spatial-/Channel-wise Attention Model ramps up the commonly used Regression CNN. The model comprises of Spatial-wise Attention Model (SAM) and Channel-wise Attention Model (CAM). 
& \cellcolor{PastePink} Density maps constructed from Point annotations & \cellcolor{PastePink} Count estimated from \par  Density Maps \\
\cellcolor{PastePink} &\cellcolor{PastePink} &\cellcolor{PastePink} &\cellcolor{PastePink} &\cellcolor{PastePink} \\

\cellcolor{PastePink}SFA-Net~\cite{zhu2019dual} &
\cellcolor{PastePink}arXiv, \par 2019 & \cellcolor{PastePink} 
 SFANet uses attention mechanism to generate high quality density maps. It uses VGG model to extract feature map followed by the application of a dual path multiscale fusion network. One of the path retrieves the attention map by extracting the crowd areas and the other path fuses the multi-scale features and attention map to generate high quality density maps.
&

\cellcolor{PastePink} Density maps constructed from Point annotations & 
\cellcolor{PastePink} Count estimated from \par Density Maps\\
\cellcolor{PastePink} &\cellcolor{PastePink} &\cellcolor{PastePink} &\cellcolor{PastePink} &\cellcolor{PastePink} \\

\bottomrule
\end{tabular}
}
\end{table*}



The skew in the data distribution and discontinuities affect all aspects of the crowd counting problem. These factors induce imbalance in minibatch sampling, optimization and evaluation. Since the default evaluation protocol (averaging over test errors) does not take the data distribution skew into account, the resulting score ({\em e.g.}, Mean Absolute Error (MAE)) exhibits high standard deviation, often $4-5$ orders of magnitude higher than the MAE itself (see Figure~\ref{fig:1}). This high deviation prevents mean scores from being considered as a reliable performance statistic. Since error deviation is not reported in literature, this issue has gone unaddressed so far.

\begin{figure*}[!t]
  \includegraphics[width=\textwidth]{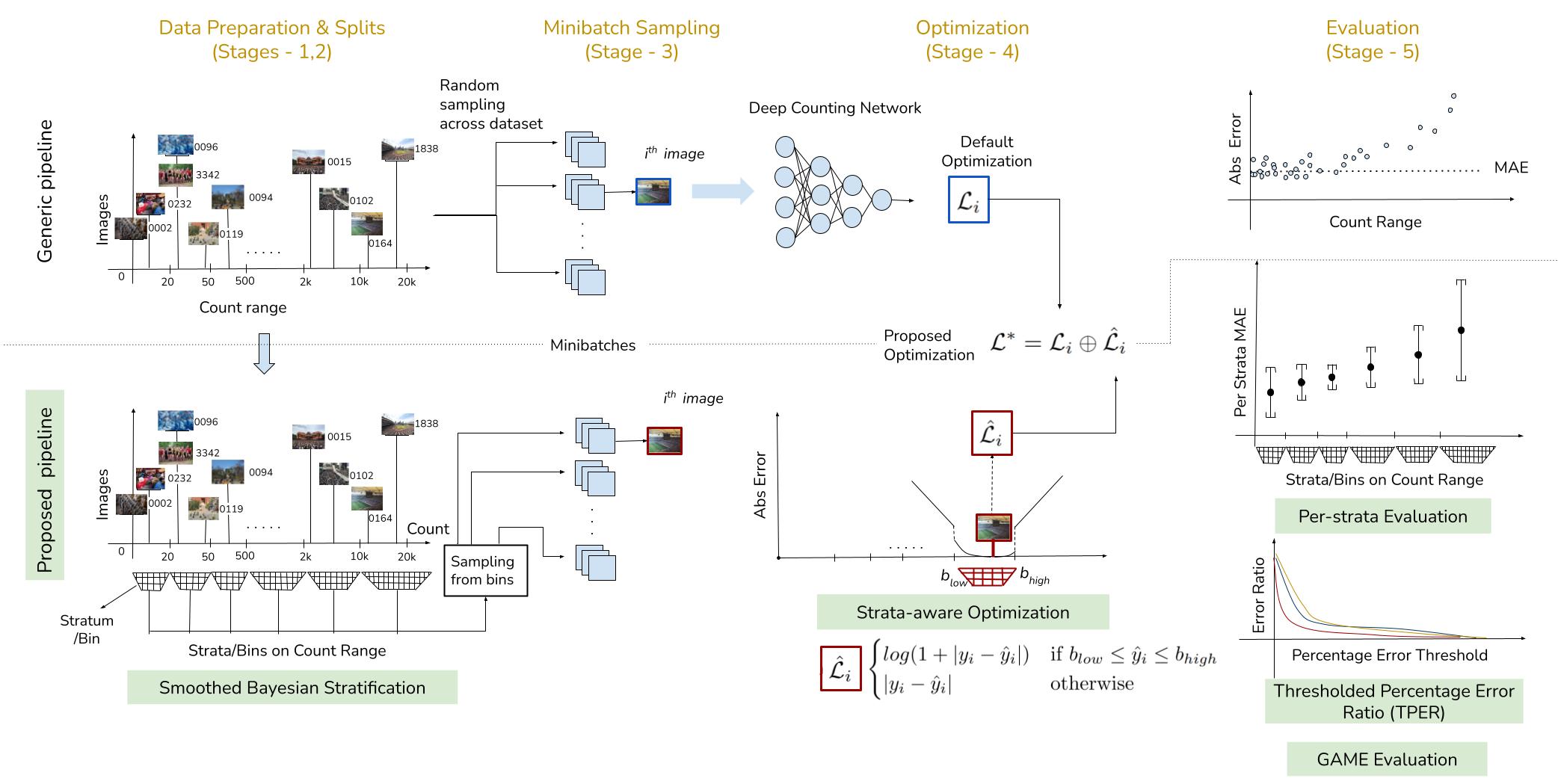}
  \caption{An overview diagram depicting the generally employed processing pipeline of a crowd-counting approach (top) and the proposed modifications we introduce in this work (bottom). See Section~\ref{sec:proposedapproach} for details.}
  \label{fig:prototypicalAndChanges}
\end{figure*}

To address the issues mentioned above, we propose an approach that actively factors in the count distribution and its skew at every stage of the problem (see Figure~\ref{fig:prototypicalAndChanges}). The contributions are two-fold, {\em viz.}, at the training stage and at the evaluation stage. During the training stage, we devise an algorithm for partitioning the count range into balanced strata (bins) using Bayesian optimality as a criterion (Sec.~\ref{sec:proposedapproach}). The balanced bins form the basis for minibatch sampling (Sec.~\ref{sec:stage3}). We also formulate a loss function that additionally penalizes error based on the ground-truth binning (Sec.~\ref{sec:stage4}).

During the evaluation, instead of reporting a single performance summary statistic (MAE) across the entire test set range, we report bin-wise statistics and aggregate these statistics in a principled manner (Sec.~\ref{subsec:stage5}) to report the overall score. Additionally, we devise a novel Thresholded Percentage Error Ratio (TPER) which builds on the relative score wrt the ground-truth count. 
This metric presents an strategy to compare across networks and across datasets. Finally, we present values on the Grid Average Mean absolute Error (GAME)~\cite{game} which factors in the error at the predicted location alongside the Absolute Error of the people count in the image. These inclusive performance measures provide deeper insights into the performance of the network and aid the end user in selecting her application specific network.

We perform comparative evaluation involving representative state-of-the-art deep counting networks~\cite{Song_2021_ICCV,Wan_2021_CVPR,liang2021focal,sasnet,wang2020DMCount,ma2019bayesian,gao2019scar,xhp2019SDCNet,zhu2019dual} and across multiple crowd counting datasets, {\em viz.}, NWPU, JHU, UCF-QNRF and ShangaiTech Part A and Part B ~\cite{gao2020nwpu,9009496,idrees2018composition,mcnn}. Our results (Sec.~\ref{sec:results}) demonstrate that the proposed approach results in a noticeable reduction of error deviation compared to the default (no-binning) procedure. Overall, our contributions help both researchers and end-users determine performance for various count ranges and select from among various approaches based on their relative performance within these ranges. Code, pretrained models and detailed visualizations can be accessed from our project page \url{deepcount.iiit.ac.in}.

\section{Related Work}

\subsection{Training Methods}
To the best of our knowledge, no works have analyzed the processing pipeline for crowd counting in entirety. In this section, we review works which aim to address some aspects raised in the earlier section. 

The crowd counting networks can be divided into set of groups based on their Ground truth-Output data type. The summary of the grouping can be seen in the Table~\ref{table:summarymodels}. The networks P2PNet~\cite{Song_2021_ICCV} (represented in purple) and FIDTM~\cite{liang2021focal}  (represented in yellow) are standalone models. The former's procedure ingests point annotations (or image pixel coordinates) during train and predicts point annotations during evaluation. The latter is a network that uses novel Focal Inverse Distance Map instead of the commonly used density map. The next group in green color consists of three networks, {\em viz.}, GenLoss~\cite{Wan_2021_CVPR}, DM-Count~\cite{wang2020DMCount} and BL~\cite{ma2019bayesian}. This group of networks takes in the point annotations as reference and predicts a density map that represents the crowd during evaluation. The last group (represented in pink) consists of networks SASNet~\cite{sasnet}, S-DCNet~\cite{xhp2019SDCNet}, SCAR~\cite{gao2019scar} and SFA-Net~\cite{zhu2019dual}. This group of networks uses the density maps as reference and generates the density maps during evaluation, from which the number of people in the image has been decided. The summary of each of these models is presented in Table~\ref{table:summarymodels}.

\subsection{Evaluation methods:} Mean Absolute Error (MAE) and Mean Squared Error (MSE) are the most prevalent evaluation measures in crowd counting approaches, with MAE usually being the more direct measure. More recently, some attempts have been made to examine MAE statistics based on percentage errors, illumination levels and scene levels to characterize performance~\cite{gao2020nwpu}. 
However, these are post-hoc measures and do not tackle imbalance which crops up in other stages of the standard pipeline employed for crowd counting. 

We introduce a novel Thresholded Percentage Error Ratio (TPER) built on Percentage Error and Error Ratio. Percentage Error is a relative metric that indicates the difference between a predicted value and target value in comparison with the target value. Error Ratio (or Rate) is a measure commonly used for classification setting and is a measure to find the ratio of correctly classified samples to the total number of samples. We extend this measure to a regression setting.

Apart from count errors, the network's ability to localize people in the image can also be of interest. P2PNet~\cite{Song_2021_ICCV} proposed density normalized Average Precision (nAP), which penalizes both localization errors and false detections. On similar lines, in the vehicle counting domain, Ricardo \textit{et al.}~\cite{game} proposed Grid Average Mean absolute Error (GAME) which simultaneously estimates the object count error and localization error.

\section{Proposed method}
\label{sec:proposedapproach}

\subsection{Standard Processing Pipeline}
\label{sec:pipe}

As depicted in Figure~\ref{fig:prototypicalAndChanges}, any standard approach to crowd-counting can be considered to have five stages:

\begin{itemize}
    \item \textit{Stage-1 (Data preparation):} In this stage, images and corresponding counts are processed suitably and are provided as input and output to a reference deep network. This stage includes standard procedures such as image cropping and resizing, density map preparation, {\em  etc.}
    \item \textit{Stage-2 (Creating data splits):} The prepared data is partitioned into training, validation and test splits according to a pre-defined split ratio ({\em e.g.}, $65\%,15\%,20\%$). 
    \item \textit{Stage-3 (Minibatch creation):} The deep network is trained using a subset of data randomly sampled from the training set, usually referred as a minibatch. The training set is partitioned into minibatches for each training epoch.
    \item \textit{Stage-4 (Optimization):} The parameters of the deep network are optimized for a loss function at the minibatch level. 
    \item \textit{Stage-5 (Evaluation):} A standard performance measure ({\em e.g.}, MAE) is used for evaluating the model on the validation or the test set. 
\end{itemize}

Each of these stages involves a set of assumptions which are often implicit. For instance, the train-validation-test splitting (Stage-2) and minibatch creation (Stage-3) assume that the distribution over the targets (counts) is uniform. However, target distributions for standard crowd counting datasets are heavy-tailed. Due to the uniform nature of sampling, the data splits and consequently, the training minibatches, exhibit the same heavy-tailed distribution. This skew induces a bias which penalizes samples in the tail during optimization (Stage-4). Due to this bias, the statistical summary measures ({\em e.g.}, MSE, MAE) fail as representative measures of performance (Stage-5). 

To address these issues, we revisit the entire problem setting and propose alternative paradigms for the stages mentioned previously. We leave Stage-1 untouched and describe our modifications to the subsequent stages. 

\subsection{Revisiting Stage 2 (Creating Data splits)}
\label{subsec:stage2}

As mentioned earlier, the standard sampling procedure for creating train-validation-test splits implicitly assumes a uniform distribution over the target range. However, doing so causes the tail portion of the distribution to be under-represented. A fundamental reason for this effect is that the sampling is conducted at too fine a resolution, {\em i.e.} at the level of individual counts. 

One approach to address this issue is to coarsen the resolution and partition the count range into bins (strata) that are optimal for uniform sampling. Formally, let the total number of images be $N$ and suppose the count range over the data samples is $R=[0,C]$, where $C$ is the maximum crowd count. The count data $\Dcal$ can be represented in terms of observed discrete counts $c_i$ and their frequencies $f_i$, as $\Dcal = \left \{ \langle c_i, f_i \rangle \left|\  i = 1,...m \right. \right\}$, where $m$ is total number of distinct counts in the dataset. Thus, $c_1=  0, c_m = C$. Consider a partitioning of the counts into $N_b$ bins as:

\begin{equation}
\label{eq:1}
    \Pscr(1,N) \equiv \lbrace [n_{k-1},n_k-1] \rbrace, k = 1,2,3\ldots N_b
\end{equation}

where $n_{k-1}$ represents the start index of the $k^{th}$ bin. Note that $n_0=0$ and $n_{N_b}-1 = C$. For simplicity, we drop the reference to $(1,N)$ when referring to $\Pscr(1,N)$ in what follows. 

\subsubsection{Partition Prior}
\label{sec:partitionprior}

We formulate the prior over partitions in terms of number of bins $N_b$ in a candidate partition. In what follows, we refer to this prior distribution as $P(N_{b})$. To avoid the degenerate case in which each unique count in the range might land up in its own bin, we impose constraints over the number of bins~\cite{Scargle2013}. Specifically, we use a geometric prior to assign lower probability to a partition containing larger bin counts:

\begin{align}
\label{eq:3}
    P(N_b;\gamma) =  
    \begin{cases}
    P_0 \; \gamma^{N_b} & \text{if } 1 \leqslant N_b \leqslant \alpha \\
    0 & \text{otherwise}
  \end{cases}
\end{align}

where $P_0$ is a normalization constant. $\gamma < 1$ is a parameter which affects the distribution profile and $\alpha$ controls the practical effectiveness of the upper bound on $N_b$. Applying the laws of probability to $P(N_b)$ and solving for $P_0$, we  obtain: 

\begin{equation}
\label{eqn:prior}
    P(N_{b};\gamma) = \frac{1 - \gamma}{1 - \gamma^{\alpha}} \gamma^{N_{b}}
\end{equation}

\subsubsection{Partition Likelihood}
\label{sec:partitionlkhood}

\begin{figure}[!t]
\centering
\includegraphics[width=0.3548\textwidth]{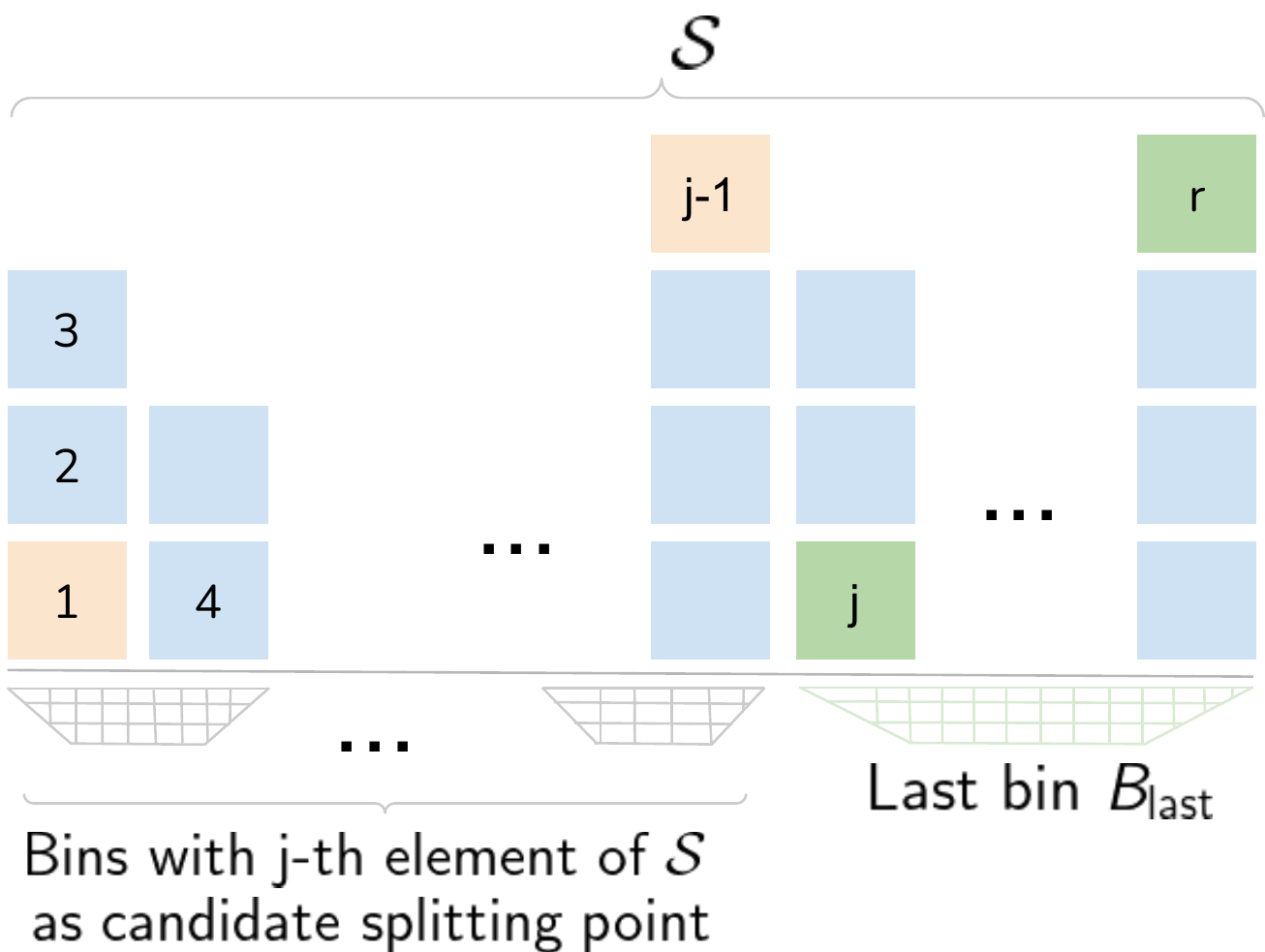}
\caption{A candidate partitioning of a subsequence of $\Scal$ ending with the $r^{th}$ element of $\Scal$. Finding the optimal partitioning can be thought of as a search over such candidate partitions. Refer to Sec.~\ref{sec:optimalpartitioning}.}
\label{fig:candidatepartition}
\end{figure}

The likelihood for a partition $\Pscr$ is defined in terms of the likelihood of each constituent bin in the partition. Let $m_k$ be the width of bin $B_k$. Let the count frequencies of the $m_k$ distinct counts within the bin be denoted by $x_1,x_2,\ldots x_{m_k}$ respectively. We model the likelihood for each bin as a multinomial distribution:

\begin{equation} \label{eqn:binlkhood}
\begin{split}
lik(B_k) & = lik(x_1,\dots , x_{m_k} ; p_1, \dots, p_{m_k}) \\
 & = \frac{X_k!}{x_1! x_2! \ldots x_{m_k}!} \prod_{j=1}^{m_k} p_{j} ^{x_j}
\end{split}
\end{equation}

where $X_k = \sum_{j=1}^{m_k} x_j$ and $p_{j}$ is probability of the $j^{th}$ count. Assuming independence across bins, the \textit{log} likelihood of the partition can be expressed as:

\begin{equation}
\label{eq:7}
    lik[\Pscr] = \sum_{k=1}^{N_b} lik(B_k)
\end{equation}

\subsubsection{Optimal Partitioning}
\label{sec:optimalpartitioning}

Given the count range $R = [0,C]$, at one extreme, we can have a partitioning wherein all data lies in a single bin. At the other extreme, we can have a partitioning wherein each unique integer in the range $R$ is a bin. Thus, finding the optimal partitioning can be thought of as a search over candidate partitions that lie between these two extremes. 

To solve this task efficiently, we adopt a dynamic programming approach~\cite{Scargle2013}. To begin with, we transform the count frequency data $\Dcal$ into a sequence of counts $c_1, c_2....c_m$ where $c_i$ is repeated $f_i$ times, {\em i.e.}, $\Scal : \lbrace c_i,c_i,\ldots (f_i $ times$, 1 \leqslant i \leqslant m) \rbrace$.  Let $\Fcal_{opt}(1,r)$ be the optimal Maximum A Potseriori (MAP) score for the partitioning of a subsequence of $\Scal$ ending with the $r^{th}$ element of $\Scal$. Following the principle of optimality, we have:

\begin{equation}
\begin{aligned}
    \Fcal_{opt}(1,r) =  
    \begin{cases}
    0, \text{ if }  r =1 \\
   \underset{1 < j \leqslant r}{\text{max}} \Big[ \texttt{ best(1,j-1)}  + lik(B_{last})(j,r) \\ \;\;\;\;\;\;\;\;\;\; + \; log \: P(b_j;\gamma) \Big] \text{ if r=2,3\ldots N} 
  \end{cases}
\end{aligned}    
\label{eqn:fopt}
\end{equation}

where \texttt{best(1,j-1)} is the memoized (precomputed and stored) best likelihood value (Eqn.~\ref{eq:7}) for the sub-partition ending in the $(j-1)^{th}$ element, $lik(B_{last})(j,r)$ is the likelihood of the final bin containing the subsequence beginning at the $\Scal$'s $j^{th}$ element and ending with the $r^{th}$ element (see Fig.~\ref{fig:candidatepartition}).  $log \: P(b_j;\gamma)$ is the prior on number of bins (Eqn.~\ref{eqn:prior}). More concretely, $b_j$ is the number of bins that form with $\Scal$'s $j^{th}$ element as the split location for the last bin.  

Note that the MAP formulation of $\Fcal_{opt}(1,r)$ incorporates the partition likelihood and prior in a Bayesian manner. With respect to the formulation in Eqn~\eqref{eqn:fopt}, the optimal set of bins corresponds to the ones obtained for $\Fcal_{opt}(1,|\Scal|)$,  where $|\Scal|$ is the number of elements in sequence $\Scal$.

\begin{algorithm}[!t]
  \caption{Optimal Bins}
  \label{algorithm:1}
  \begin{algorithmic}[1]
  \Procedure{OptimalBins}{$\Dcal$}
\LineComment{\textbf{Input} data $\Dcal$ }
\LineComment{\textbf{Output} Optimal bins $bins_{best}$}
\LineComment{\textcolor{blue}{Grid search values for $\gamma$ (Sec.~\ref{sec:partitionprior})}}
 \State $\Gamma = [0.1,0.2,\ldots0.9]$
\LineComment{\textcolor{blue}{Grid search values for train-test ratios}}
 \State $ratios = [0.1,0.2,0.25]$
\LineComment{\textcolor{blue}{Cross-validation repeat factor}}
 \State $seeds= 10$

 \For {$\gamma \text{ in } \Gamma$}
     \For{$r \text{ in } ratios$}
         \For{$f \text{ in } [0:1:seeds]$}
                \State $\Dcal_{f}$ = shuffle($\Dcal$,$seed=f$);
                \LineComment{\textcolor{blue}{Algorithm~\ref{algorithm:2}}}
                \State $lik_{f,r,\gamma}$ = \Call{FindLikelihood}{$\Dcal_{f},r,\gamma$}
         \EndFor
        \LineComment{\textcolor{blue}{Compute average likelihood for a fixed $\gamma$ and $r$}}
         \State $lik_{r,\gamma} =$ \Call{Mean}{$lik_{f,r,\gamma}$}
    \EndFor
\EndFor
\LineComment{\textcolor{blue}{To find the best $\gamma$ across all $ratios$,}}
\LineComment{\textcolor{blue}{descending sort by likelihood for each ratio $r$.}}
\LineComment{\textcolor{blue}{For each $\gamma$, sum indices of corresponding location}}
\LineComment{\textcolor{blue}{in sorted order of earlier step.}}
\For {$\gamma \text{ in } \Gamma$}
\State $idxsum_{\gamma} = 0$
\For{$r \text{ in } ratios$}
\State $idxsum_{\gamma} += $ \Call{GetDescendingIndx}{$lik_{r,\gamma}$}
\EndFor
\EndFor
\LineComment{\textcolor{blue}{The best $\gamma$ is one with lowest index sum.}}
\State $\gamma_{best}$ = $\underset{\gamma}{\text{ argmin }} idxsum_{\gamma}$
\LineComment{\textcolor{blue}{Use the best $\gamma$ and determine optimal partitions (Sec.~\ref{subsec:stage2}).}}
\State $bins_{best}$ =  \Call{BayesianOptimalBins}{$\Dcal,prior=\gamma_{best}$}
  \EndProcedure
\end{algorithmic}  
\end{algorithm}

\begin{algorithm}[!t]
\caption{Algorithm to find likelihood of a held out subset}
\begin{algorithmic}[2]
\Procedure{FindLikelihood}{$\Dcal,ratio,\gamma$}
\LineComment{\textbf{Input} Data $\Dcal$,train-test split ratio $ratio$, prior param $\gamma$ }
\LineComment{\textbf{Output} Likelihood $lik$ of $\Dcal$'s test subset}
\LineComment{\textcolor{blue}{Split data into train, test as per $ratio$}}
\State $train$ $,test$ $=$ \Call{SplitData}{$\Dcal,ratio$}
\LineComment{\textcolor{blue}{Find optimal bins using train set} (Sec.~\ref{subsec:stage2})} 
\State $bins=$ \Call{BayesianOptimalBins}{$train,prior=\gamma $}
\LineComment{\textcolor{blue}{Find likelihood of test set}}
\LineComment{\textcolor{blue}{wrt optimal bins found earlier (Sec.~\ref{sec:partitionlkhood})}}
\State $lik$ $=$ \Call{ComputeBinsLkhood}{$test,bins$}
\EndProcedure
\end{algorithmic}
\label{algorithm:2}
\end{algorithm}

\subsubsection{Additive Smoothing}
\label{sec:additivesmoothing}

The sample distribution in crowd datasets is not only heavy tailed, but also sparse at the tail end. In other words, the distribution is characterized by large count spans which do not have any sample associated with them. This causes the binning procedure described in this section to output a large number of sparsely filled bins. To mitigate this effect, we perform additive smoothing~\cite{daniel} on the data before binning. Formally, a smoothing factor $\beta$ is added to each distinct count across the count range $R=[0,C]$. In our case, $\beta=1$.

\subsubsection{Grid-search for optimal hyper-parameters}
\label{sub:modbay}

To determine the optimal set of bins, we first perform a grid search with cross-validation over a range of values for (i) distribution profile parameter $\gamma$ (Eqn.~\ref{eq:3}) (ii) the train-validation split ratios. Having determined the optimal hyper-parameter $\gamma_{best}$, we utilize the same to obtain the optimal set of bins, as outlined in Algorithm ~\ref{algorithm:1}. 
\subsection{Revisiting Stage 3: Minibatch Creation}
\label{sec:stage3}

To address the skew induced by the heavy-tailed, discontinuous count distribution of data samples, we bin the data optimally using the procedure described in Section~\ref{subsec:stage2}. To populate a minibatch using our Round Robin (RR) method, we pick a data sample randomly from each of the bins in a round robin fashion, beginning at the first bin. This process is repeated until all the bins have been selected or the minibatch is full. We continue this process until the entire training dataset is accounted for as an epoch ({\em i.e.}, in terms of minibatches). This procedure is followed for each epoch. 

Another variant of binning which we consider is Random Sampling (RS) procedure where a bin is first picked randomly from available bins and a data sample is picked randomly from the randomly selected bin. A procedure similar to Round Robin (RR) is used to populate an epoch's equivalent of training data. Effectively, both our procedures ensure that the mini-batches are balanced in terms of their count range unlike the standard random shuffle-based approach. We analyze the results with both the binning strategies as part of our evaluations (Sec.~\ref{sec:results}).

\subsection{Revisiting Stage 4: Optimization}
\label{sec:stage4}

The standard protocol for optimizing a deep counting network is to minimize the per-instance loss averaged over the minibatch. However, one is confronted with the same issues (imbalance, bias) as those faced during minibatch creation (Sec.~\ref{sec:stage3}). As a consequence, the trained networks exhibit high variance for the error term $|y - \hat{y}|$, where $y$ is the ground-truth count and $\hat{y}$ is the predicted count.

\begin{figure}[!t]
\centering
\includegraphics[width=0.45\textwidth]{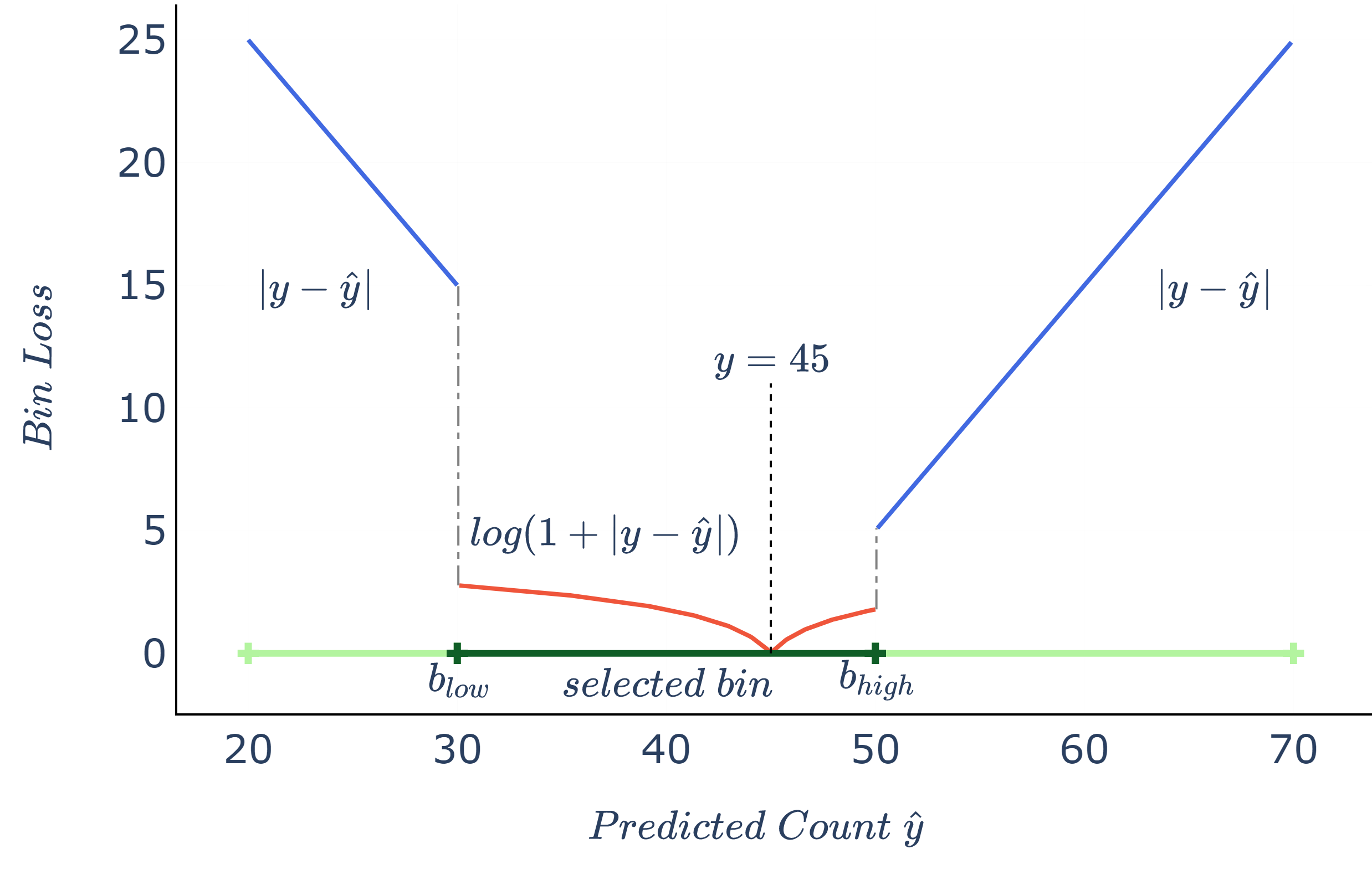}
\caption{Bin Loss Function : The figure depicts the ground truth count $y=45$ and the loss function variation with respect to the predicted count $\hat{y}$ inside the bin ($log(1+ |y - \hat{y}|)$) and outside ($ |y - \hat{y}|$). The reference bin is highlighted in dark green. Refer to Sec.~\ref{sec:stage4} for details.}
\label{fig:3}
\end{figure}

To enable data-distribution aware optimization, we introduce a novel bin sensitive loss function $\widehat{\mathcal{L}}$. Instead of the loss depending solely on the error, we also consider the count bin to which the data sample belongs and whether the predicted count $\hat{y}$ lies within this bin or outside it. If $\hat{y}$ lies within the bin, we impose a smaller logarithmic penalty.  If the count value lies outside, we impose a linear penalty. Formally, our strata-aware loss function is defined as:  

\begin{equation}
  \widehat{\mathcal{L}} =   \begin{cases}\label{eq:8}
  \lambda_1 \; log(1+ |y - \hat{y}|) & \text{if $ b_{low} \leqslant  \hat{y}  \leqslant b_{high} $ } \\
  |y - \hat{y}| & \text{otherwise}
\end{cases} 
\end{equation}

where $b_{low}$ and $b_{high}$ are defined by the bin that $y$ belongs to (see Fig.~\ref{fig:3}) and $\lambda_1$ is a weighting factor of the log component. This loss is added as an additive component to the default model loss to encourage strata-aware optimization.

\subsection{Revisiting Stage 5: Evaluation}
\label{subsec:stage5}

The discontinuous and heavy-tailed distribution of samples affects the evaluation stage as well. Coupled with lack of bin-level awareness during optimization, an outlier effect surfaces, which causes the default measures ({\em e.g.}, MSE, MAE) to be ineffective representatives of performance \textit{across} the entire count range. Even more worryingly, the standard deviation of error tends to be in the order of 4-5 times the mean statistic. Instead of using a single pair of numbers (mean, standard deviation) to characterize performance across the entire count range, we make the following proposals.

\textit{One,} the evaluation measure must be reported at the level of each bin. This provides a more comprehensive picture of performance. Additionally, it also helps compare the relative effectiveness of various counting networks for smaller and larger counts.

\textit{Two,} even if an overall summary statistic over the test set is deemed necessary, the mean and standard deviation of bin-level performance measures are combined in a statistically sound manner. Let the mean and standard deviations for the individual bins be $(\mu_i,\sigma_i); i=1,2,\ldots N_b$ and let the number of samples in each bin be $n_i$. We compute the pooled mean and standard deviation as their weighted average:
    
\begin{equation}
    \mu_{pool} =\frac{n_1 \mu_1 +  n_2 \mu_2 + \ldots + n_{N_b} \mu_{N_b} }{n_1 + n_2 + \ldots + n_{N_b}}
\end{equation}

\begin{equation}
    \sigma^2_{pool} =\frac{n_1 \sigma^2_1 +  n_2 \sigma^2_2 + \ldots + n_{N_b} \sigma^2_{N_b} }{n_1 + n_2 + \ldots + n_{N_b}} 
\end{equation}

\begin{figure}[!t]
\centering
\includegraphics[width=0.5\textwidth]{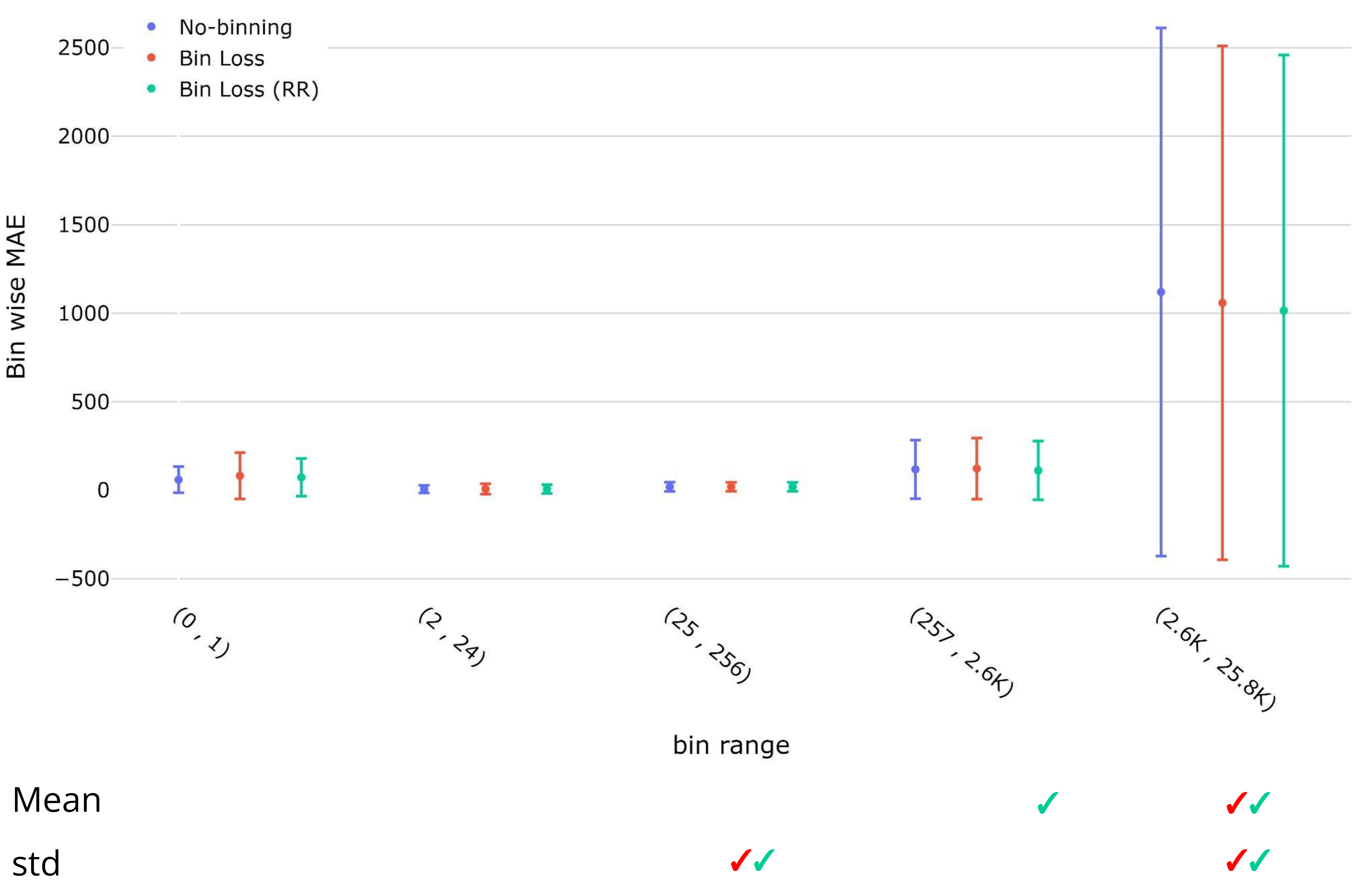}
\caption{Per-bin performance of BL on JHU dataset. The ticks present below the Per-bin performance plots indicate the bins in which the Bin Loss (RS) (in \textcolor{red}{red}) and Bin Loss (RR) (in \textcolor{NewGreen}{green}) perform better than the No-binning setting wrt to Mean and std respectively. In this plot, binning reduces the mean and standard deviation in larger count range bins 25 to 25.8k.}
\label{fig:bin-wise-jhu}
\end{figure}

\begin{figure}[!t]
\centering
\includegraphics[width=0.5\textwidth]{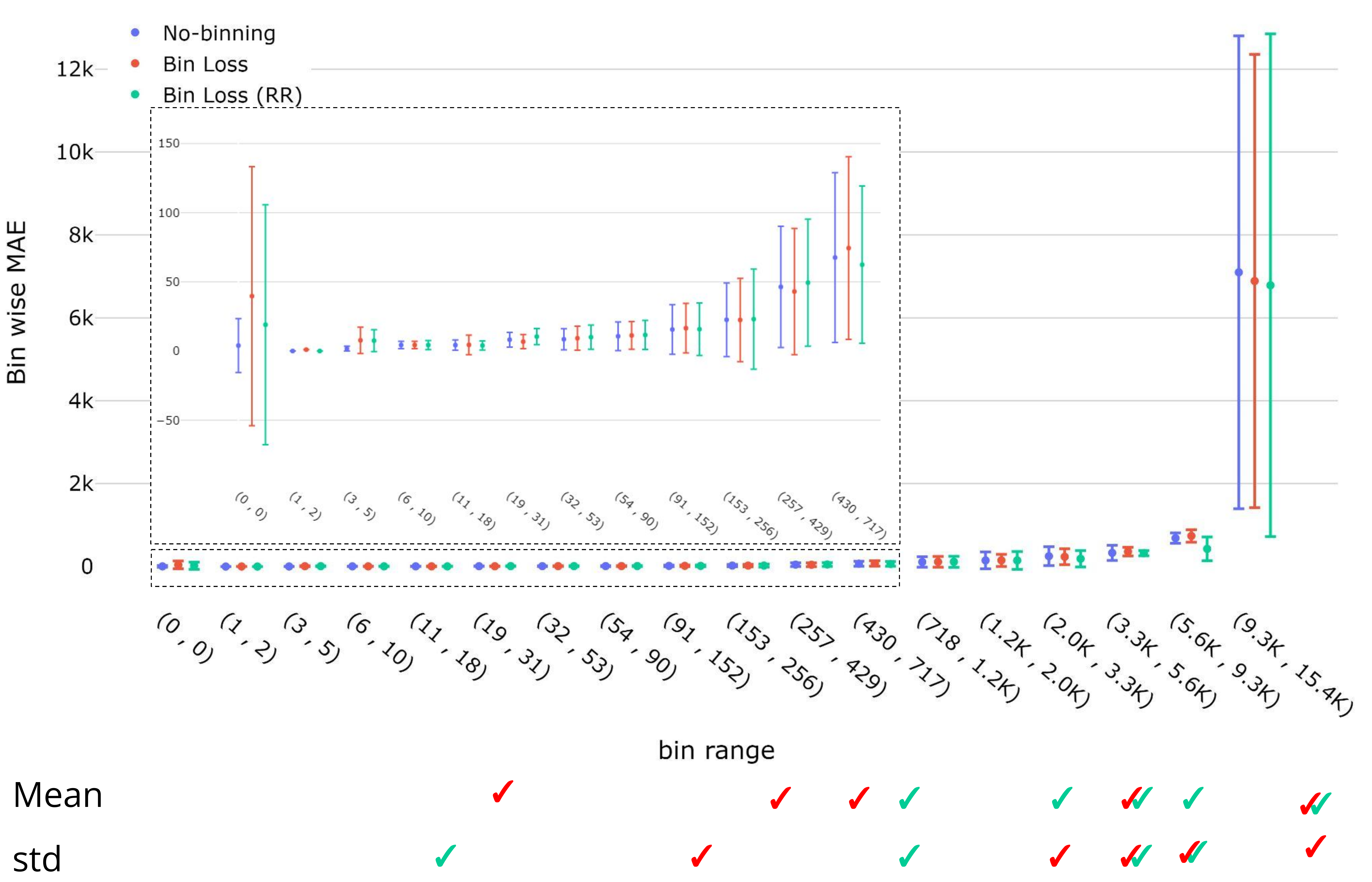}
\caption{Per-bin performance of FIDTM on NWPU dataset. Similar to our observation in the earlier plot, the comparatively larger deviations for the no-binning scheme are clearly evident.}
\label{fig:bin-wise-nwpu}
\end{figure}

 \textit{Three,} a relative error metric must be reported {\em i.e.} significance of the error is wrt the ground-truth count. The customarily used relative performance measure Mean Absolute Percentage Error (MAPE) is not a ideal metric in this scenario using the same rationale as the MAE (Sec~\ref{section:intro}). As proxy to include relativity, we propose a novel Thresholded Percentage Error Ratio ($TPER$). The formulation for TPER is derived from Percentage Error ($PE$) and Error Ratio. We compute the $PE$ using the ground-truth count ($y$) and  predicted count($\hat{y}$):
 
\begin{equation}
\label{eqn:pe}
    PE =\frac{ |y - \hat{y}| }{ y } 
\end{equation}

For the formulation of TPER, we fix $\theta$'s as linearly increasing values between $[0,100]$ in steps of 5. The TPER over the test set ($\Dcal_{test}$ with $T$ samples) is defined by:

\begin{equation}
\label{eqn:tper}
    TPER_{\theta} =\frac{\text{ \# of images for which } PE >= \theta }{T} 
\end{equation}

Combining equations Eq.~\ref{eqn:pe} and Eq.~\ref{eqn:tper}, we formulate the final form of $TPER$:

\begin{equation}
    TPER_{\theta} =\frac{\text{ \# of images for which } ( |y - \hat{y}| >= \theta * y )}{T} 
\end{equation}

\begin{figure}[!t]
\centering
\includegraphics[width=0.5\textwidth]{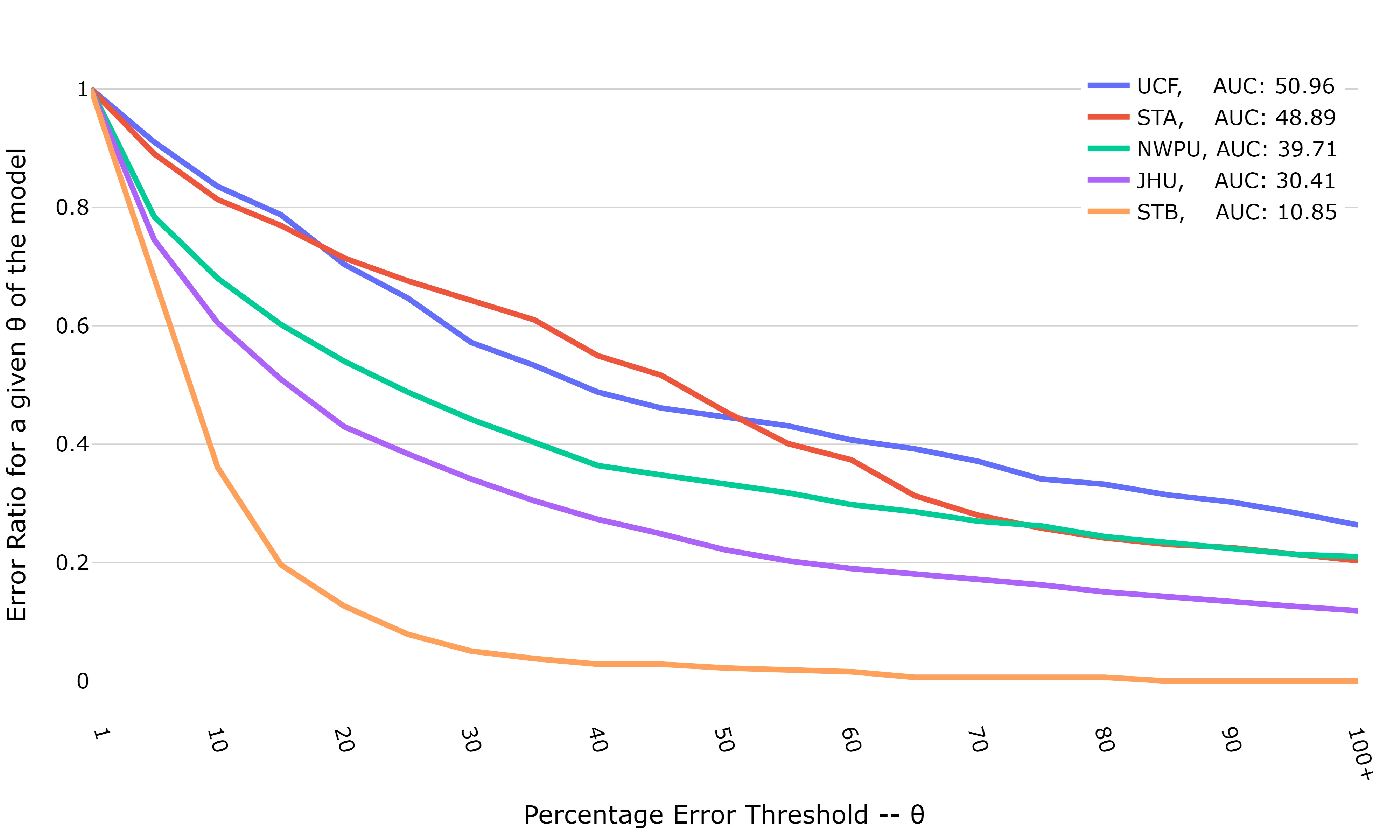}
\caption{Plot between percentage Error threshold (x-axis) and TPER (y-axis) for the network BL~\cite{ma2019bayesian} across ~\cite{9009496,gao2020nwpu,idrees2018composition,mcnn} datasets. This plot symbolizes the robustness of the network over different datasets.}
\label{fig:singlemodel}
\end{figure}

\begin{figure}[!t]
\centering
\includegraphics[width=0.5\textwidth]{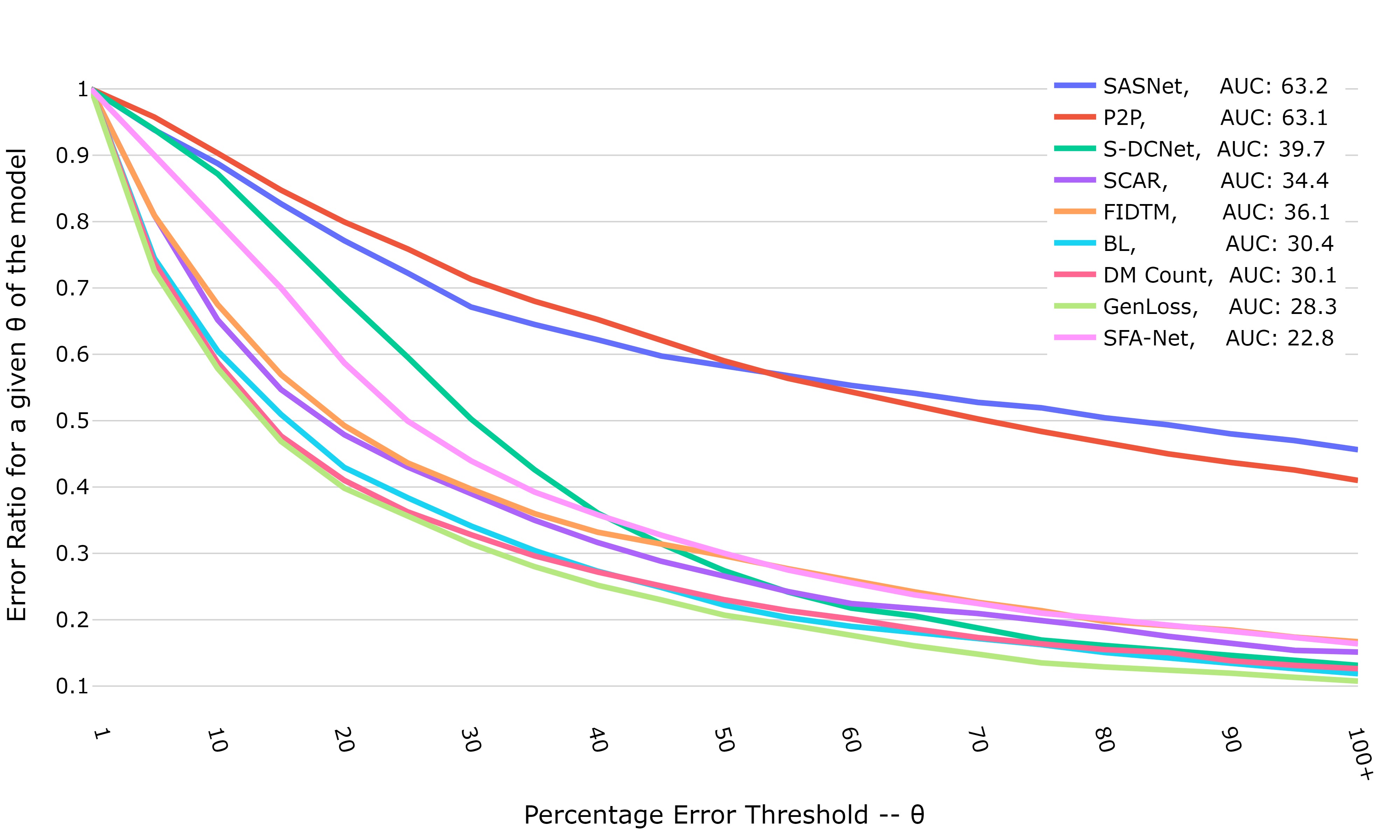}
\caption{Plot of Percentage Error Threshold (x-axis) and TPER (y-axis) for the different models ~\cite{Song_2021_ICCV,Wan_2021_CVPR,liang2021focal,sasnet,wang2020DMCount,ma2019bayesian,gao2019scar,xhp2019SDCNet,zhu2019dual} over a single dataset JHU~\cite{9009496}. Most  methods converge to lower error rate other than P2PNet~\cite{Song_2021_ICCV} and SASNet~\cite{sasnet}.}
\label{fig:singledataset}
\end{figure}

For a given threshold $\theta$, $TPER_{\theta}$ represents the percentage of test images that have Absolute Error ($|y - \hat{y}|$) greater than $\theta$ times the ground-truth count $y$. We graphically represent the trend of the TPER wrt to the increasing $\theta$'s. Furthermore, the Area Under the Curve (AUC) is indicated in the Figure~\ref{fig:singlemodel} and ~\ref{fig:singledataset}. The AUC is a  numerical quantification about the performance based on the TPER. 

This type of evaluation can be utilized to compare the network's robustness across different datasets as shown in Figure~\ref{fig:singlemodel}. TPER also provides a proxy percentage error profile for the best performing network for a given dataset Figure~\ref{fig:singledataset}.

\begin{figure}[!t]
\centering
\includegraphics[width=0.5\textwidth]{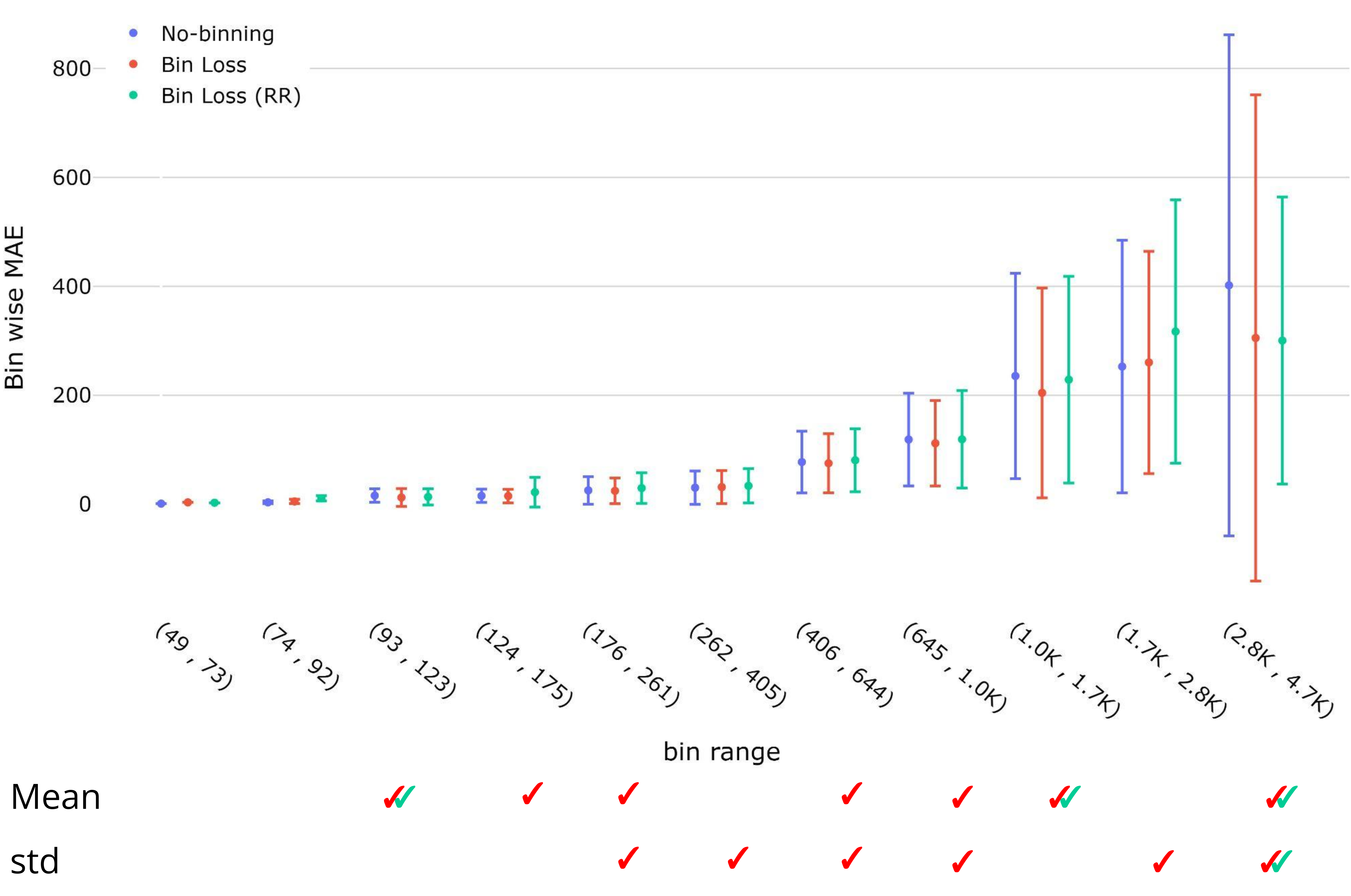}
\caption{Per-bin performance of GenLoss on UCF dataset. The Bin Loss (Random Sampling - RS) approach out-performs the No-binning in 9 out of 11 bins.}
\label{fig:bin-wise-ucf}
\end{figure}

\begin{figure}[!t]
\centering
\includegraphics[width=0.5\textwidth]{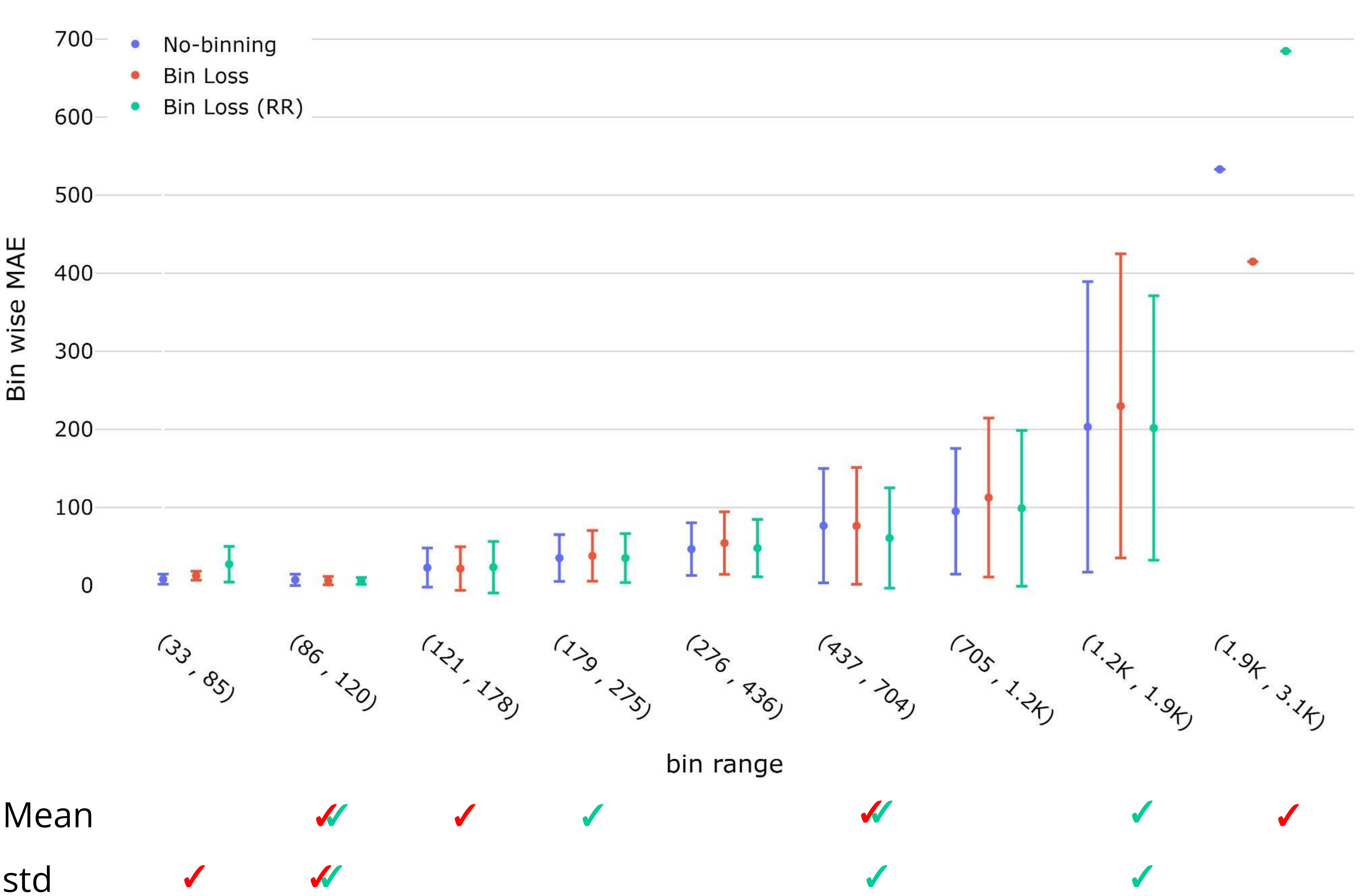}
\caption{Per-bin performance of S-DCNet on STA dataset. Introducing binning improves mean and standard deviation in more than half of the cases (indicated by tick marks)}
\label{fig:bin-wise-sta}
\end{figure}

\begin{figure}[!t]
\centering
\includegraphics[width=0.5\textwidth]{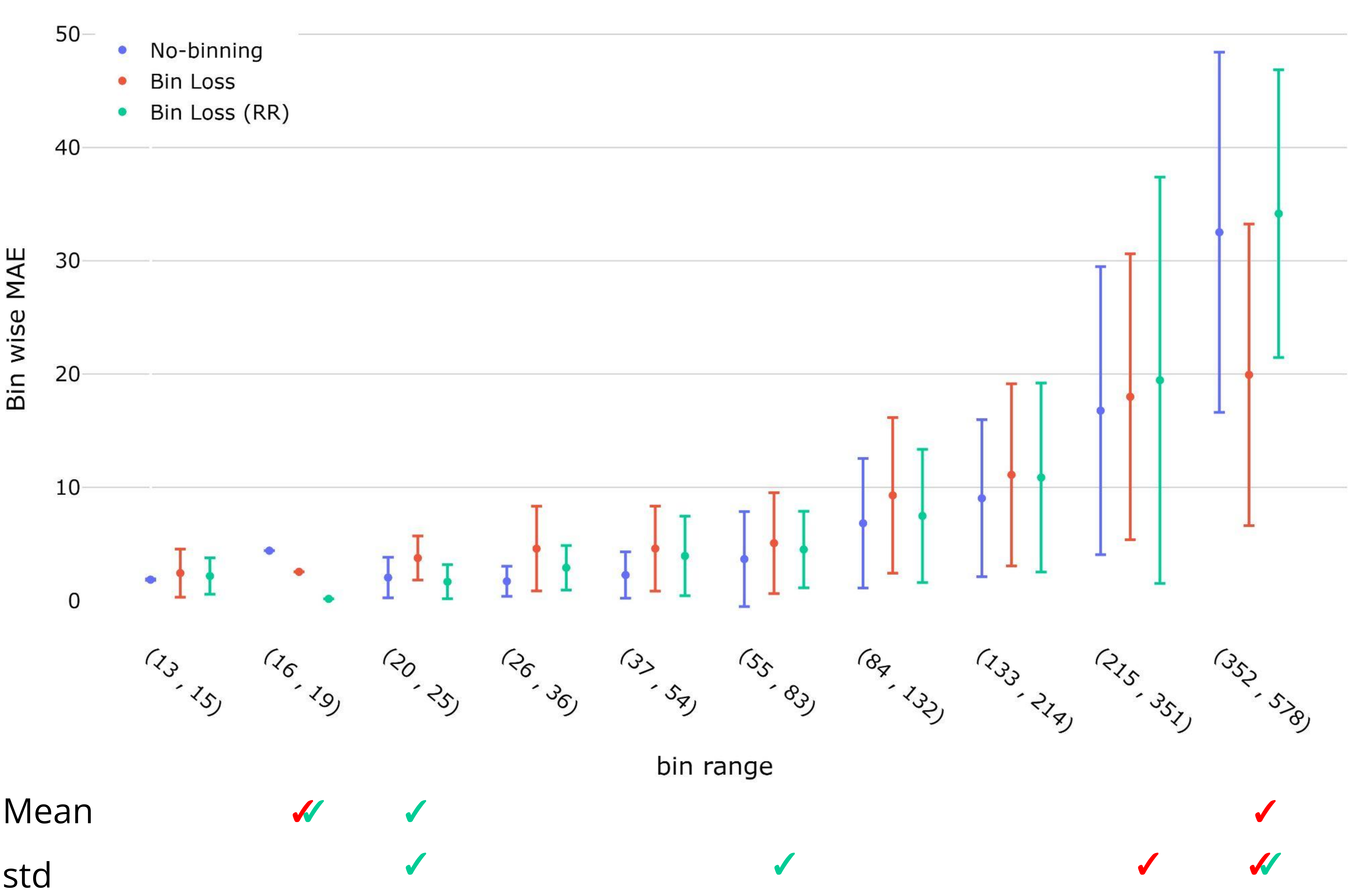}
\caption{Per-bin performance of SFA-Net on STB dataset. Relatively lower standard deviation is observed in the cases of Bin Loss and Bin Loss (RR).}
\label{fig:bin-wise-stb}
\end{figure}

\textit{Four,} 
apart from the drawbacks of Mean Absolute Error(MAE) mentioned in Section~\ref{section:intro}, MAE evaluation conveniently skips the measurement of localization performance. Localization metrics in counting are important to keep in check the false positives and locations of predicted heads.

To provide localized evaluation, GAME~\cite{game} is a metric which penalizes both the error in estimated location of the head along and the error in the predicted count.  With GAME~\cite{game}, we subdivide the image into $4^{L}$ non-overlapping regions, where $L \in \mathbb{Z}$, and calculate localized MAE within each  subregions. Mathematically,

\begin{equation}
    GAME(L) =\frac{1}{T}\sum_{t=1}^{T}(\sum_{l=1}^{4^L} |y_t^l - \hat{y}_t^l|) 
\end{equation}
where, $\hat{y}_t^l$ and $y_t^l$ are the predicted count and the ground-truth count in a region $l$ for the image $t \in T$.

Observably, as $L$ increases, the number of sub-regions increase exponentially, leading to smaller evaluated regions and higher GAME error. Similar trend can be observed in Table~\ref{table:gametable}. 

\begin{table*}[!t]

\caption{Evaluation results on four benchmark datasets JHU, NWPU, UCF-QNRF, ShanghaiTech-A,B (STA,STB)  using the evaluation procedure in Sec.~\ref{subsec:stage5} on diverse models. The size of test set is indicated below dataset name. The columns represent minibatching schemes (Bin Loss(RS): random bin selection , Bin Loss(RR): round robin bin selection, No-binning: default procedure without binning). For each result, superscript denotes the standard deviation. The best result for each dataset is highlighted in sky blue. The group-wise best is indicated by its group color. The best MAE and standard deviation of the absolute errors are highlighted in bold for each network. Note that first three columns of the table (Pooled MAE and standard deviation) are not directly comparable to the last column of Global MAE and standard deviation values.}
\label{table:results}
\centering
\resizebox{0.9\linewidth}{!}
{
\centering

\begin{tabular}{c|c|r|ccc|| c}
\toprule
& & & \multicolumn{3}{c||}{Pooled MAE and std} & \multicolumn{1}{c}{Global MAE and std} \\
\midrule
Size of Dataset & Dataset & Model  & Bin loss (RS) & Bin loss (RR) & No-binning & No-binning \\

\toprule

\multirow{24}{*}[-0.25em]{\rule{0pt}{2ex} Large }
& \multirow{8}{*}[-0.25em]{\rule{0pt}{2ex} $\text{\normalsize{JHU}} \cite{9009496} \atop \mathbf{1600}$}
& \cellcolor{PasteLavender} P2PNet \cite{Song_2021_ICCV}   & $204.4^{\pm \mathbf{318.2} }$ & \cellcolor{PasteLavender} $\mathbf{193.2}^{\pm 327.6 }$ & $211.5^{\pm 365.7 }$ & $211.5 ^{\pm 538.1 }$\\

& & \cellcolor{PasteYellow} FIDTM \cite{liang2021focal} &  \cellcolor{PasteYellow} $\;\:\mathbf{71.2}^{\pm \mathbf{208.5} }$ & $\;\:73.0^{\pm 210.1 }$ & $\;\:89.3^{\pm 278.7 }$ & $\;\:89.3^{\pm 323.6 }$  \\

& & \cellcolor{PasteGreen} GenLoss \cite{Wan_2021_CVPR}   & $\;\:63.5^{\pm 215.7 }$ & $\;\;\:61.7^{\pm \mathbf{211.8} }$ & $\:\mathbf{61.4}^{\pm 217.5 }$ & $\;\:61.4 ^{\pm 261.1 }$\\
& & \cellcolor{PasteGreen} DM-Count \cite{wang2020DMCount} & $\;\:72.0^{\pm 224.0 }$ & $\;\:69.6^{\pm 213.3 }$ & $\;\;\mathbf{62.1}^{\pm \mathbf{207.4} }$ & $\;\:62.1 ^{\pm 244.0 }$ \\
& & \cellcolor{PasteGreen} BL \cite{ma2019bayesian}   & $\;\:64.3^{\pm 213.3 }$ & \cellcolor{SkyBlue}$\;\:\mathbf{60.6}^{\pm \mathbf{210.7} }$ & $\;\:63.9^{\pm 216.4 }$ &  $\;\:63.9 ^{\pm 262.1 }$  \\

& & \cellcolor{PastePink} SASNet \cite{sasnet} & $\!\!\mathbf{289.9}^{\pm 117.6 }$ & $292.3^{\pm 117.6 }$ & $\;294.7^{\pm \mathbf{117.4} }$ & $294.7 ^{\pm 716.5 }$\\
& & \cellcolor{PastePink}  S-DCNet \cite{xhp2019SDCNet} & $\;\:90.2^{\pm 225.9 }$ & $\;\mathbf{86.7}^{\pm 227.9 }$ & $\;\;\:88.1^{\pm \mathbf{199.5} }$ & $\;\:88.1 ^{\pm 313.6 }$  \\
& & \cellcolor{PastePink}  SCAR \cite{gao2019scar} & $\;\;\, 78.9^{\pm \mathbf{218.9 }}$ & $\;\:79.1^{\pm 225.2 }$ & $\;\,\mathbf{78.7}^{\pm 223.4 }$ & $\;\, 78.7 ^{\pm 283.2 }$   \\
& & \cellcolor{PastePink} SFA-Net \cite{zhu2019dual} & \cellcolor{PastePink} $\: \mathbf{78.6}^{\pm 255.4 }$ & $\;\:82.7^{\pm 250.1 }$ & $\;\;\;82.5^{\pm \mathbf{234.8} }$ & $\;\,82.5 ^{\pm 277.1 }$   \\

\cmidrule(lr){2-7}

& \multirow{8}{*}[-0.25em]{\rule{0pt}{2ex} ${ \text{\normalsize{NWPU}}  \cite{gao2020nwpu} \atop \mathbf{500}}$  } 
& \cellcolor{PasteLavender} P2PNet \cite{Song_2021_ICCV}  & $379.0^{\pm \mathbf{131.4} }$ & \cellcolor{PasteLavender} $\mathbf{339.8}^{\pm 218.2 }$ & $366.0^{\pm 160.8 }$ & $ 366.0 ^{\pm 955.4 }$ \\

& & \cellcolor{PasteYellow} FIDTM \cite{liang2021focal} & $\;\;\:72.0^{\pm \mathbf{351.5} }$ &  \cellcolor{SkyBlue}$\;\mathbf{67.4}^{\pm 390.0 }$ & $\;\:69.2^{\pm 366.4 }$ &  $\;\:69.2 ^{\pm 580.8}$   \\

& &  \cellcolor{PasteGreen} GenLoss \cite{Wan_2021_CVPR}  & $110.9^{\pm 322.8 }$ & $ 107.3^{\pm 331.6 }$ & $ \mathbf{105.2}^{\pm \mathbf{291.6} }$ & $105.2^{\pm541.7 }$ \\
& & \cellcolor{PasteGreen} DM-Count \cite{wang2020DMCount} & $\;\:88.1^{\pm 236.7 } $ & \cellcolor{PasteGreen} $\;\:\mathbf{76.7}^{\pm \mathbf{205.0}}$ & $\;\;77.8^{\pm 214.9 }$ & $\;\;\;\;\;\; \; 71.7 ^{\pm376.4}$ \cite{wang2020DMCount}  \\
& & \cellcolor{PasteGreen} BL \cite{ma2019bayesian}  & $112.9^{\pm 333.7 }$ & $\: 114.8^{\pm \mathbf{320.3} }$ & $\!\mathbf{102.5}^{\pm 348.2 }$ & $ 102.5 ^{\pm 560.6 } $   \\

& & \cellcolor{PastePink} SASNet \cite{sasnet}  & $355.1^{\pm 116.0 }$ & $359.5^{\pm 116.0 }$ & $\mathbf{341.5}^{\pm \mathbf{115.3} }$ & $ 341.5 ^{\pm 954.9 }$ \\
& & \cellcolor{PastePink} S-DCNet \cite{xhp2019SDCNet} &  $213.4^{\pm 231.0}$  & $\: 224.1^{\pm \mathbf{230.1} }$  & $\! \mathbf{210.0}^{\pm 303.1 }$ & $\;\,248.7 ^{\pm 1161.9 } $ \\
& & \cellcolor{PastePink} SCAR \cite{gao2019scar}  & $\:112.8^{\pm \mathbf{321.3}}$ & $111.9^{\pm 325.6 }$  & \cellcolor{PastePink}  $\!\mathbf{111.3}^{\pm 332.1 }$ & $ 111.3 ^{\pm 555.8 } $  \\
& & \cellcolor{PastePink} SFA-Net \cite{zhu2019dual}  & $136.0^{\pm 299.1 }$ & $\mathbf{116.4}^{\pm \mathbf{285.2} }$ & $\, 125.0^{\pm 343.0 }$  & $\;\: 163.4 ^{\pm 1072.1 } $ \\

\cmidrule(lr){2-7}

& \multirow{8}{*}[-0.25em]{\rule{0pt}{2ex} $\text{\normalsize{UCF}} \cite{idrees2018composition}\atop \mathbf{334}$} 
& \cellcolor{PasteLavender} P2PNet \cite{Song_2021_ICCV}  & \cellcolor{PasteLavender} $\mathbf{535.7}^{\pm 254.5 }$ & $579.5^{\pm \mathbf{206.6} }$ & $599.3^{\pm 264.8 }$ &  $ 599.3 ^{\pm 637.0 }$\\

& & \cellcolor{PasteYellow} FIDTM \cite{liang2021focal} & \cellcolor{PasteYellow} $\!\, \mathbf{109.7}^{\pm \mathbf{112.4} }$ & $120.4^{\pm 150.3 }$ & $215.2^{\pm 372.4 }$ & $215.2^{\pm 432.5 }$ \\

& & \cellcolor{PasteGreen} GenLoss \cite{Wan_2021_CVPR}  & \cellcolor{SkyBlue}$\;\mathbf{89.5}^{\pm 115.4 }$ & $\; 100.1^{\pm \mathbf{109.0} }$ & $\;\;95.7^{\pm 121.0 }$ & $\;\:95.7 ^{\pm 152.5 }$ \\
& & \cellcolor{PasteGreen} DM-Count \cite{wang2020DMCount} & $\:103.8^{\pm \mathbf{107.5} }$ & $\;\;97.9^{\pm 109.1 }$ & $\;\mathbf{94.5}^{\pm 111.6 }$ &  $\;\;\;\;\;\;\:85.9^{\pm 120.6} $ \cite{wang2020DMCount} \\
& & \cellcolor{PasteGreen} BL \cite{ma2019bayesian}  & $\;\:\mathbf{91.1}^{\pm \mathbf{100.3 }}$ & $\;\;92.1^{\pm 105.8 }$ & $\;\;98.3^{\pm 134.2 }$  &  $\;\;\;\;\;\;\:87.1^{\pm 126.8} $ \cite{ma2019bayesian} \\

& & \cellcolor{PastePink} SASNet \cite{sasnet}  & $663.2^{\pm 146.8 }$ & $\mathbf{631.9}^{\pm \mathbf{145.6} }$ & $668.5^{\pm 145.8 }$ & $ 668.5 ^{\pm 745.7 }$ \\
& & \cellcolor{PastePink} S-DCNet \cite{xhp2019SDCNet} & $\:205.9^{\pm \mathbf{157.8} }$ & \!$\!\mathbf{199.2}^{\pm 164.8 }$ & $215.2^{\pm 190.0 }$ &  $ 214.7 ^{\pm 277.7 } $  \\
& & \cellcolor{PastePink} SCAR \cite{gao2019scar}  & $\:124.5^{\pm \mathbf{128.6} }$  & \cellcolor{PastePink} \!$\!\mathbf{122.9}^{\pm 129.0 }$ & $123.4^{\pm 146.9 }$ &  $ 123.4 ^{\pm 197.1 } $ \\
& & \cellcolor{PastePink} SFA-Net \cite{zhu2019dual}  & $\! \mathbf{128.6}^{\pm \mathbf{133.4} }$ & $128.9^{\pm 162.9 }$ &  $128.7^{\pm 163.2 }$ &  $ 128.7 ^{\pm 199.9 } $  \\

\midrule

\multirow{16}{*}[-0.25em]{\rule{0pt}{2ex} Medium }
& \multirow{8}{*}[-0.25em]{\rule{0pt}{2ex} $ \text{\normalsize{STA}}  \cite{mcnn}   \atop \mathbf{182} $ } 
& \cellcolor{PasteLavender} P2PNet \cite{Song_2021_ICCV}   & $67.0^{\pm 65.5 }$ & $67.3^{\pm 54.4 }$ & \cellcolor{PasteLavender} $\mathbf{63.1}^{\pm \mathbf{54.3 }}$ &  $\;\;\;\;\;\;\;58.3 ^{\pm 76.7 }$ \cite{Song_2021_ICCV}\\

& & \cellcolor{PasteYellow} FIDTM \cite{liang2021focal} & \cellcolor{PasteYellow} $\! \mathbf{66.3}^{\pm 76.6 }$ & $71.3^{\pm 84.9 }$ & $\;68.9^{\pm \mathbf{76.4} }$ &  $68.9^{\pm 96.8 }$ \\

& & \cellcolor{PasteGreen} GenLoss \cite{Wan_2021_CVPR}  & $76.5^{\pm 73.8 }$ & $69.6^{\pm 72.2 }$ & $\mathbf{68.9}^{\pm \mathbf{68.3} }$ &  $68.9 ^{\pm 93.3 }$ \\
& & \cellcolor{PasteGreen} DM-Count \cite{wang2020DMCount} & $\mathbf{88.6}^{\pm \mathbf{64.4 }}$ & $89.6^{\pm 75.9 }$ &  $\, 93.0^{\pm 81.3 }$  &  $\;\;\;\;\:\:64.1^{\pm 78.4} $ \cite{wang2020DMCount}\\
& & \cellcolor{PasteGreen} BL \cite{ma2019bayesian}  & \cellcolor{PasteGreen} $\!\mathbf{68.6}^{\pm 69.9 }$ & $68.9^{\pm 63.3 }$  & $\: 68.7^{\pm \mathbf{61.9} }$  &  $\;\;\;\;\;\, 63.5 ^{\pm 74.7 }$ \cite{ma2019bayesian}\\

& & \cellcolor{PastePink} SASNet \cite{sasnet} & $\! \!\! \!\mathbf{365.2}^{\pm 70.2 }$ & $\!\!\!379.9^{\pm 68.9 }$ & $\!419.6^{\pm \mathbf{66.1} }$ &  $419.6 ^{\pm 352.5 }$ \\
& & \cellcolor{PastePink} S-DCNet \cite{xhp2019SDCNet} & $66.6^{\pm 72.6 }$ & \cellcolor{SkyBlue}$\mathbf{60.5}^{\pm \mathbf{65.5} }$ & $\, 61.3^{\pm 66.9 }$  &  $\, 61.3 ^{\pm 88.7 } $   \\
& & \cellcolor{PastePink} SCAR \cite{gao2019scar}  & $83.7^{\pm 67.4 }$ & $\mathbf{72.9}^{\pm \mathbf{61.8} }$ & $79.3^{\pm 67.4 }$ &  $\, 79.3 ^{\pm 82.9 } $   \\
& & \cellcolor{PastePink} SFA-Net \cite{zhu2019dual}  & $68.4^{\pm 65.1 }$ & $64.9^{\pm 59.5 }$ & $\mathbf{63.6}^{\pm \mathbf{55.6} }$  &  $\, 63.6 ^{\pm 92.9 } $ \\

\cmidrule(lr){2-7}

& \multirow{8}{*}[-0.25em]{\rule{0pt}{2ex} $\text{\normalsize{STB}} \cite{mcnn} \atop \mathbf{316}$} 
& \cellcolor{PasteLavender} P2PNet \cite{Song_2021_ICCV}   & $12.0^{\pm 12.9 }$ & $8.8^{\pm 9.0 }$ & \cellcolor{PasteLavender} $\mathbf{7.6}^{\pm \mathbf{8.3} }$ &  $\;\:7.6^{\pm 10.3}$ \\

& & \cellcolor{PasteYellow} FIDTM \cite{liang2021focal} & $\, 8.7^{\pm \mathbf{8.6} }$ & \cellcolor{PasteYellow} $\mathbf{7.8}^{\pm 9.3 } $ & $\;\:9.1^{\pm 10.8 }$ &  $\;\:9.5^{\pm 14.1 }$ \\

& & \cellcolor{PasteGreen} GenLoss \cite{Wan_2021_CVPR} & \cellcolor{PasteGreen} $\mathbf{8.2}^{\pm \mathbf{9.6} }$ & $\;\:9.4^{\pm 11.9 }$ & $\;\:8.8^{\pm 11.4 }$ &  $\;\:8.8 ^{\pm 13.9 }$ \\
& & \cellcolor{PasteGreen} DM-Count \cite{wang2020DMCount} & $9.1^{\pm 9.3 }$ & $\,\mathbf{8.6}^{\pm \mathbf{8.6} }$ & $\;\;8.9^{\pm 10.3 }$  &  $\;\;\;\;\;7.3^{\pm 9.3} $~\cite{wang2020DMCount}\\
& & \cellcolor{PasteGreen} BL \cite{ma2019bayesian}  & $\! \mathbf{9.6}^{\pm 9.3 }$ & $\,9.7^{\pm 9.3 }$ & $\!\!\,10.8^{\pm \mathbf{9.2} }$  &  $\;\;\;\;\;7.5^{\pm 9.4}$~\cite{ma2019bayesian} \\

& & \cellcolor{PastePink} SASNet \cite{sasnet} & $\!\!\,\mathbf{66.8}^{\pm 21.2 }$ & $\,67.2^{\pm 21.1 }$ & $\:67.4^{\pm \mathbf{20.6} }$ &  $67.4 ^{\pm 67.8 }$ \\
& & \cellcolor{PastePink} S-DCNet \cite{xhp2019SDCNet} & $9.2^{\pm 9.4 }$ & $\;\:9.6^{\pm 10.5 }$ & $\:\mathbf{7.9}^{\pm \mathbf{8.6} }$  &  $\;\:7.8 ^{\pm 11.0 } $ \\
& & \cellcolor{PastePink} SCAR \cite{gao2019scar} & $\;\: \mathbf{9.8}^{\pm \mathbf{10.2 } }$ & $\!\:13.8^{\pm 11.7 }$ & $\,{10.3}^{\pm 14.0 }$ & $10.3 ^{\pm 14.1 } $ \\
& & \cellcolor{PastePink} SFA-Net \cite{zhu2019dual} & $\: 9.0^{\pm 7.3 }$ & $\,8.8^{\pm 8.0 }$ & \cellcolor{SkyBlue}$\,\mathbf{7.4}^{\pm \mathbf{6.8} }$  & $\,7.4 ^{\pm 9.2 } $  \\

\bottomrule
\end{tabular}
}

\end{table*}

\begin{table*}[!t]

\caption{Evaluation results in the form of GAME values over four benchmark datasets JHU, NWPU, UCF-QNRF, ShanghaiTech-A,B (STA,STB). The top 3 performing networks are shown in the table below. The columns represent minibatching schemes (Bin Loss(RS): random bin selection , Bin Loss(RR): round robin bin selection, NB: default procedure without binning). The best performer between these three minibatching schemes is bolded (mean and std) network-wise. Group-wise best is indicated by the groups color. }
\label{table:gametable}
\centering
\resizebox{0.9\linewidth}{!}
{
\centering

\begin{tabular}{c|r|ccc|ccc|ccc}
\toprule
& & \multicolumn{9}{c}{GAME metric and std\; } \\
\toprule
Dataset & Model  & \multicolumn{3}{c}{L=1} & \multicolumn{3}{c}{L=2} &\multicolumn{3}{c}{L=3} \\
\cmidrule(lr){3-11}
 &   & RS & RR & NB & RS & RR & NB & RS & RR & NB \\

\toprule

\multirow{3}{*}[-0.25em]{\rule{0pt}{2ex} $\text{\normalsize{JHU}} \cite{9009496} \atop \mathbf{1600}$}

& \cellcolor{PasteGreen} GenLoss \cite{Wan_2021_CVPR}    
& $77.1 ^{\pm 274.7}$ & $76.2 ^{\pm 269.0}$ & \cellcolor{PasteGreen} $\mathbf{75.8} ^{\pm \mathbf{271.8}}$ 
& $\:\:\, 89.4 ^{\pm 289.7}$ & $\;\;\; 87.9 ^{\pm \mathbf{283.1}}$ & \cellcolor{PasteGreen} $\mathbf{87.6} ^{\pm 285.2}$ 
& $107.2 ^{\pm 299.9}$ & $\, 105.3 ^{\pm \mathbf{292.4}}$ & \cellcolor{PasteGreen} $\!\!\!\! \mathbf{105.0} ^{\pm 294.2}$\\

& \cellcolor{PasteGreen} DM-Count \cite{wang2020DMCount}  
& $86.9 ^{\pm 292.4}$ & $85.2  ^{\pm 290.2}$ & $\; \mathbf{80.0} ^{\pm \mathbf{270.3}}$ 
& $\:\:\, 98.0 ^{\pm 305.2}$ & $\:\:\, 96.6 ^{\pm 301.8}$ & $\:\:\, \mathbf{91.5} ^{\pm \mathbf{281.3}}$
&  $114.8 ^{\pm 313.5}$  & $113.3 ^{\pm 310.6}$ & $\mathbf{108.3} ^{\pm \mathbf{291.3}}$ \\

& \cellcolor{PasteGreen} BL \cite{ma2019bayesian}    
& $78.4 ^{\pm 270.4}$ & $\mathbf{76.7} ^{\pm \mathbf{265.3}}$ & $\; 80.7 ^{\pm 281.7}$ 
& $\:\:\, 91.2 ^{\pm 283.7}$ & $\:\:\, \mathbf{89.3} ^{\pm \mathbf{279.8}}$ & $\:\:\, 92.3 ^{\pm 292.1}$
& $110.6 ^{\pm 294.6}$ & $\!\, \mathbf{107.9} ^{\pm \mathbf{289.9}}  $ & $111.1 ^{\pm 301.9}  $ \\

\midrule

\multirow{3}{*}[-0.25em]{\rule{0pt}{2ex} ${ \text{\normalsize{NWPU}}  \cite{gao2020nwpu} \atop \mathbf{500}}$  }

& \cellcolor{PasteYellow} FIDTM \cite{liang2021focal} 
& $\, 80.5 ^{\pm \mathbf{561.9}}$ & $77.3 ^{\pm 580.3}$ &\cellcolor{PasteYellow}  $\mathbf{76.2} ^{\pm 581.2}$
& $\;\;\; 88.2 ^{\pm \mathbf{562.0}}$ & $\;\; 85.6 ^{\pm 581.0}$ & \cellcolor{PasteYellow} $\mathbf{82.1} ^{\pm 581.7}$
& $\, 101.2 ^{\pm \mathbf{563.8}}$ & $\;\; 98.6 ^{\pm 583.4}$ & \cellcolor{PasteYellow} $\mathbf{92.4} ^{\pm 582.9}  $ \\

& \cellcolor{PasteGreen} GenLoss \cite{Wan_2021_CVPR}    
& $\!\!\!\!\, 117.3  ^{\pm 629.6}$ & $\!\! 116.3 ^{\pm 623.5}$ & $\mathbf{114.7} ^{\pm \mathbf{546.1}}$ 
& $128.2 ^{\pm 630.0}$ & $128.2 ^{\pm 633.7} $ & $\mathbf{127.8} ^{\pm \mathbf{575.8}}$
& $\!\! \mathbf{149.2} ^{\pm 635.4}$ & $149.7 ^{\pm 644.9}$ & $\; 149.6 ^{\pm \mathbf{597.2}}$ \\

& \cellcolor{PasteGreen} DM-Count \cite{wang2020DMCount}  
& $\!\, 94.1  ^{\pm 508.6}$ & \cellcolor{PasteGreen} $\!\!\! \mathbf{82.9} ^{\pm 437.1}$ & $\;\;\; 83.7 ^{\pm \mathbf{392.2}}$
& $102.5 ^{\pm 510.6}$ & \cellcolor{PasteGreen} $\mathbf{90.8} ^{\pm 439.0}$ & $\;\;\; 91.5 ^{\pm \mathbf{398.2}} $
& $116.8 ^{\pm 516.7}$ & \cellcolor{PasteGreen} $\!\!\!\! \mathbf{105.1} ^{\pm 445.1}$ & $\; 105.9 ^{\pm \mathbf{404.4}}$ \\

\midrule

\multirow{3}{*}[-0.25em]{\rule{0pt}{2ex} $\text{\normalsize{UCF}} \cite{idrees2018composition}\atop \mathbf{334}$}

& \cellcolor{PasteYellow} FIDTM \cite{liang2021focal}
& $\! 267.9 ^{\pm 959.9}\ $ & \cellcolor{PasteYellow} $\!\!\!\!\! \mathbf{145.2} ^{\pm \mathbf{192.7}}$ & $245.1 ^{\pm 433.9}$
& $291.3 ^{\pm 964.6}$ & \cellcolor{PasteYellow} $\!\! \mathbf{170.8} ^{\pm \mathbf{213.6}}$ & $\, 272.4 ^{\pm 445.6}$
& $320.1 ^{\pm 965.2}$ & \cellcolor{PasteYellow} $\!\!\!\!\, \mathbf{203.9} ^{\pm \mathbf{236.5}}$ & $303.4 ^{\pm 455.0}$ \\

& \cellcolor{PasteGreen} GenLoss \cite{Wan_2021_CVPR}    
& \cellcolor{PasteGreen} $\!\!\!\!\! \mathbf{101.1} ^{\pm \mathbf{145.1}}$ & $\!\!\! 110.5 ^{\pm 146.5}$ & $106.3 ^{\pm 155.8}$
& \cellcolor{PasteGreen} $\!\!\! \mathbf{118.5} ^{\pm 151.3}$ & $\: 127.2 ^{\pm \mathbf{151.1}}$ & $\, 124.9 ^{\pm 162.0}$
& \cellcolor{PasteGreen} $\!\!\!\! \mathbf{145.4} ^{\pm 167.0}$ & $\, 154.0 ^{\pm \mathbf{166.5}}$ &  $153.9 ^{\pm 178.3}  $ \\

& \cellcolor{PasteGreen} BL \cite{ma2019bayesian} 
& $\!\!\, 105.1 ^{\pm \mathbf{136.8}}$ & $\!\!\!\! \mathbf{104.5} ^{\pm 140.0}$ & $112.9 ^{\pm 167.8}$ 
& $124.0 ^{\pm 152.8}$ & $\mathbf{121.5} ^{\pm \mathbf{144.1}} $ & $\, 132.8 ^{\pm 176.6}$
& $152.6 ^{\pm 168.5}  $ & $\! \mathbf{149.7} ^{\pm \mathbf{157.0}}$ & $162.9 ^{\pm 190.4} $ \\

\midrule
\multirow{3}{*}[-0.25em]{\rule{0pt}{2ex} $ \text{\normalsize{STA}}  \cite{mcnn}   \atop \mathbf{182} $ } 

& \cellcolor{PasteYellow} FIDTM \cite{liang2021focal} 
& \cellcolor{PasteYellow} $\!\!\! \mathbf{78.5} ^{\pm 104.9}$ &  $81.2 ^{\pm 108.7}$ & $\;\;\; 82.2 ^{\pm \mathbf{101.7}}$
& \cellcolor{PasteYellow} $\mathbf{91.0} ^{\pm 104.2} $ & $\;\; 93.4 ^{\pm 108.8}$ &  $\;\;\, 97.0 ^{\pm \mathbf{103.4}}$
& $\! 117.1 ^{\pm 111.3}$ & \cellcolor{PasteYellow} $\!\!\!\!\! \mathbf{116.8} ^{\pm 113.8}$ &  $\, 123.2 ^{\pm \mathbf{108.9}}$ \\

& \cellcolor{PasteGreen} GenLoss \cite{Wan_2021_CVPR}   
& $ \!\!\! 84.4 ^{\pm 94.7}$ & $\!\! 78.0 ^{\pm 94.5}$ & $\mathbf{76.7} ^{\pm \mathbf{93.5}}$
& $97.4 ^{\pm 96.3}$ & $91.0 ^{\pm 95.1}$ & $\mathbf{89.6} ^{\pm \mathbf{94.8}}$ 
& $\!\!\!\! 122.8 ^{\pm 99.9}$ & $\!\!\!\! 118.1 ^{\pm 99.8}$ & \cellcolor{PasteGreen} $\!\!\!\!\! \mathbf{116.1} ^{\pm \mathbf{97.5}}$ \\

& \cellcolor{PasteGreen} BL \cite{ma2019bayesian}    
& $\!\!\! 77.1 ^{\pm 87.0}$ & $\!\! 76.2 ^{\pm 92.3}$ & \cellcolor{PasteGreen} $\!\! \mathbf{75.5} ^{\pm \mathbf{85.2}}$
& $\, 91.2 ^{\pm \mathbf{88.1}} $ & $90.5 ^{\pm 93.4} $ & \cellcolor{PasteGreen} $\!\!\! \mathbf{89.2} ^{\pm 88.7}$
& $\!\!\!\! 121.4 ^{\pm 95.1}  $  & $\!\!\!\! 119.7 ^{\pm 98.3} $ & $\!\! \!\mathbf{119.0} ^{\pm \mathbf{93.8}} $ \\

\midrule

\multirow{3}{*}[-0.25em]{\rule{0pt}{2ex} $\text{\normalsize{STB}} \cite{mcnn} \atop \mathbf{316}$} 

& \cellcolor{PasteYellow} FIDTM \cite{liang2021focal}
& \cellcolor{PasteYellow} $\!\!\!\!\! \mathbf{11.6} ^{\pm 12.1}$ & $\! 11.7 ^{\pm \mathbf{11.9}}$ & $12.9 ^{\pm 14.7}$ 
& \cellcolor{PasteYellow} $ \!\! \mathbf{15.9} ^{\pm \mathbf{13.5}}$ & $16.6 ^{\pm 14.1}$ & $\, 16.9 ^{\pm 15.7}$
& \cellcolor{PasteYellow} $\!\!\! \mathbf{22.5} ^{\pm \mathbf{16.8}}$ & $24.0 ^{\pm 17.0}$ & $24.1^{\pm 19.3}$
\\

& \cellcolor{PasteGreen} GenLoss \cite{Wan_2021_CVPR}    
& \cellcolor{PasteGreen} $\!\!\!\!\! \mathbf{10.4} ^{\pm \mathbf{12.5}}$ & $\!\! 11.7 ^{\pm 14.9}$ & $10.7 ^{\pm 14.1}$
& $\: 15.0 ^{\pm \mathbf{14.1}}$ & $16.1 ^{\pm 16.4}$ & \cellcolor{PasteGreen} $\!\! \mathbf{14.8} ^{\pm 14.9}$
& $\, 23.7 ^{\pm \mathbf{17.8}}$ & $24.6 ^{\pm 19.9}$ & \cellcolor{PasteGreen} $\!\!\! \mathbf{22.8} ^{\pm 18.0}$ 
\\

& \cellcolor{PasteGreen} DM-Count \cite{wang2020DMCount}  
& $\!\!\! 11.5 ^{\pm 11.8}$ & $\!\!\!\! \mathbf{11.1} ^{\pm 11.2}$ & $11.7 ^{\pm 12.9}$
& $\, 15.9 ^{\pm 13.0}$ & $\mathbf{15.4} ^{\pm \mathbf{12.6}}$ & $\, 15.8 ^{\pm 13.7}$
& $23.8 ^{\pm 16.3}$ & $\mathbf{23.0} ^{\pm \mathbf{16.0}}$ & $\, 23.3 ^{\pm 16.6} $ 
\\

\bottomrule
\end{tabular}
}

\end{table*}

\section{Experimental Setup}
\label{sec:experiments}

We perform experiments with three large-scale crowd counting datasets JHU~\cite{9009496}, NWPU~\cite{gao2020nwpu} and UCF-QNRF~\cite{idrees2018composition}  as well as two variants of the medium-scale dataset ShanghaiTech(A,B)~\cite{mcnn}. Although we revisit all stages of the problem pipeline, we retain the standard train and test datasets for consistency. To determine optimal bin hyper-parameters (Section~\ref{subsec:stage2}), we isolate a random $20\%$ subset of the train set and use the same for validation. Since NWPU's test set is not directly available, we use the publicly available validation set as the test set and report results on the same. We also compare the two different binning schemes mentioned in Section~\ref{sec:stage3}, {\it viz.}, round-robin (RR) and random selection (RS).
For evaluation, we utilize representative and recent state-of-the-art crowd counting networks, {\em viz.}, P2PNet~\cite{Song_2021_ICCV}, FIDTM~\cite{liang2021focal}, Generalised Loss (GenLoss)~\cite{Wan_2021_CVPR}, DM-Count~\cite{wang2020DMCount}, Bayesian Crowd Counting (BL)~\cite{ma2019bayesian}, SASNet~\cite{sasnet}, SCAR~\cite{gao2019scar}, SFA-Net~\cite{zhu2019dual}, S-DCNet~\cite{xhp2019SDCNet}. These papers report results on the ShanghaiTech  and UCF-QNRF datasets but not on heavier datasets like NWPU or JHU (except for FIDTM, GenLoss, DM-Count). Therefore, we report respective test set results by training these networks on the NWPU dataset and JHU dataset as well. 

The network architecture, ground truth generation, augmentation and image pre-processing steps are used as mentioned in the respective works. We employ the hyperparameters, optimizers and loss functions used as suggested in the original implementations of the networks. As mentioned previously, we add the bin-aware loss function (Sec.~\ref{sec:stage4}) to the original loss function used by the models during optimization. We compute the per-bin MAE and associated standard deviation. We also aggregate the resulting statistics to obtain an overall performance score across the bins (Sec.~\ref{subsec:stage5}). Although not directly comparable to our proposed performance score, we also report the standard MAE (which does not involve any binning) as computed by existing works. As a new addition, we also report the error's standard deviation. For baseline comparison, we also train models using the default (no-binning) procedure and without the bin-aware loss function included. We also report a novel Thresholded Percentage Error Ratio (TPER), which is used to compare performance across datasets and across network. Finally, we report GAME~\cite{game} metric for $ L  \in \{1,2,3\}$  which indicates the localization performance for top 3 performing networks in each dataset.

\section{Results}
\label{sec:results}

\subsection{Bin-level results}

The bin-level mean error scores and the corresponding standard deviation bars can be viewed for a selection of different datasets and binning schemes in Figures~\ref{fig:bin-wise-jhu},~\ref{fig:bin-wise-nwpu},~\ref{fig:bin-wise-ucf},~\ref{fig:bin-wise-sta} and~\ref{fig:bin-wise-stb}. The comparatively large deviations typically incurred when binning is not used can clearly be seen. Also note that the bin-level plots provide a larger perspective on the performance of the approach across the count range, in contrast to a single number which is conventionally reported. We have indicated every bin (right below the respective bin) for which the binning (RR in green, RS in red) yields lower bin-level mean error and standard deviation than the No-binning setting.

\subsection{Aggregate results}

The aggregate scores (described in Section~\ref{subsec:stage5}) can be viewed in Table (Pooled MAE and std) ~\ref{table:results}. Across networks and datasets, a reduction in error standard deviation is clearly apparent when bin-aware loss is used (relative to the no-binning counterpart). The aggregate scores reinforce the trend seen in the bin-level plots discussed previously. The reduction in standard deviation compensates for the marginally inferior mean score (compared to no-binning) in some cases. In the Table~\ref{table:results}, the group wise best is indicated by its group color (purple, yellow, green and pink) and the best network for a given dataset is indicated in blue. Based on the group-wise observations we can note that :

\begin{itemize}
    \item The purely point based approaches (purple) benefits from binning in the case of large datasets.
    \item The Focal Inverse distance Map approach (yellow) makes best use of binning by performing better than the no binning case in all the datasets.
    \item Approaches that ingest point annotations and predict density maps (green) give their best performances using binning over all the datasets.
    \item The approaches that deal with density maps at ground-truth and output stages (pink) take advantage of binning two-out-of-three times in large datasets and one-out-of-two times in medium datasets.
\end{itemize}

Blue highlighted results in Table~\ref{table:results} indicate, binning schemes provide the best overall aggregate results across the datasets (except for the smaller count STB dataset). Our project page \url{deepcount.iiit.ac.in} contains interactive visualizations for examining results on a per-dataset and per-model (approach) basis.

In the last column of Table (Global MAE and std)~\ref{table:results}, we also present the usually reported MAE measure. The results using models made available by authors are indicated. For the first time, we also report the standard deviation for the sake of completeness and consistency. Note that the numbers in this column are not directly comparable with other columns of the table due to the significant differences across the processing pipeline stages. However, the magnitude of the deviation incurred even by the state of the art approaches is somewhat alarming. It is also interesting to note that the Global MAE performance ranking for different networks differs significantly from the binning (Pooled MAE) results. For instance, GenLoss~\cite{Wan_2021_CVPR} is the best performer on UCF with Pooled MAE, but the same does not hold for Global MAE. A similar trend can be seen for the JHU, STA and STB datasets as well.

In our experiments, we tried two minibatching schemes (balanced, random) to determine their effect on performance, if any (Section \ref{sec:stage3}). The aggregate results across datasets suggests that random sampling and round robin sampling have better overall performance approximately half the time (Table~\ref{table:results}, first two columns). Also, the results suggest that round robin sampling of bins works best for most of the datasets (JHU, NWPU, STA). 

\subsection{Thresholded Percentage Error Ratio (TPER)}
In our experiments we represent the TPER values as curves wrt to the varying thresholds $\theta$ as shown in Figures~\ref{fig:singlemodel} and ~\ref{fig:singledataset}. The TPER indicates the ratio of test images that have the error greater than $\theta$ times the ground-truth count. As the threshold increases, the Error Ratio decreases. The slope by which the curve declines represents the performance of the network. A steeper curve (lower Area Under the Curve - AUC) indicates lesser images with high relative error and vice versa.

The Figure~\ref{fig:singlemodel} indicates the robustness or stability of the network over various datasets. If the network was dataset invariant, the curves would be clustered together. Apart from the smaller dataset STB, we can observe that there are almost 15\% of test samples in UCF, STA, NWPU and JHU datasets that have Error more than 100 times the ground-truth count for the network BL~\cite{ma2019bayesian}. The next Figure~\ref{fig:singledataset}, compares different networks on the JHU dataset. The networks SASNet~\cite{sasnet} and P2PNet~\cite{Song_2021_ICCV} have more than 40\% images with Percentage Error Threshold ($\theta$) 100. Rest of the networks have less than 20\% of the test samples falling in the same range of $\theta$.

\subsection{Grid Average Mean absolute Error (GAME)~\cite{game}}

In our experiments, we use GAME~\cite{game} as a localization measure. 
This metric effectively measures both predicted count error and its location.
We have experimented with varying $L$ values for the datasets (JHU, NWPU, UCF, STA, STB) and the networks (P2PNet, FIDTM, GenLoss, DM-Count, BL, SASNet, S-DCNet, SCAR, SFA-Net) , out of which the top 3 are mentioned in the Table~\ref{table:gametable}. 

Focal Inverse Distance Map Approach (yellow) is among the top 3 performers in NWPU, UCF, STA and STB. Approaches that ingest point annotations and predict density maps (in green) remains on the top for small as well as large datasets. Overall, the localization performance is improved when Binning (RS and RR) is introduced. There is a performance dip in the group of networks that use density maps for both ground-truth and output (represented in pink). This group performed reasonably well when evaluated by MAE or overall count, but fails when localization performance is considered.

\subsection{Ablation Studies}
\label{sec:abl}

For ablation studies, we conducted experiments with DM-Count~\cite{wang2020DMCount} on NWPU dataset. The loss function involved in optimization (Sec.~\ref{sec:stage4}) is of the form 

\subsubsection{Ablations on  $\lambda_1,\lambda_2$}
\begin{equation}
\mathcal{L}^* = \mathcal{L}  + \lambda_2 \widehat{\mathcal{L}} 
\end{equation}

where $\mathcal{L}^*$ is the final loss function, $\mathcal{L}$ is the model loss, $\widehat{\mathcal{L}}$ is the Bin Loss and $\lambda_2$ is a weighting factor. From Eqn.~\ref{eq:8}, we need to tune for both $\lambda_1,\lambda_2$. We conduct a grid optimization with $\lambda_2$ ranging over $\{0.01,1\}$ and $\lambda_1$ over $\{1,10,100\}$. The pooled MAE and standard deviations are summarized in Table~\ref{table:hyp}. Based on the results, we fix $\lambda_1= 1,\lambda_2= 1$ for DM-Count~\cite{wang2020DMCount} on all datasets and minibatching schemes (RR,RS).
\begin{table}[!t]
\caption{Hyperparameter search for $\lambda_1$ and $\lambda_2$ over a grid and the resulting pooled MAE and standard deviations.}
\label{table:hyp}
\centering
\resizebox{0.75\linewidth}{!}
{
\centering
\begin{tabular}{c|c|c}
\toprule
     $\lambda_1 \downarrow$  $\lambda_2 \rightarrow$ & 0.01 & 1 \\
\toprule
1 & $84.1^{\pm 183.2 }$ & $76.7^{\pm 205.0 }$ \\
10 & $ 80.5^{\pm 243.7 }$ & $79.5^{\pm 238.7 }$ \\
100 & $ 80.7^{\pm 236.8 } $  & $80.4^{\pm 252.7 }$\\
\bottomrule
\end{tabular}
}
\end{table}

The effectiveness of bin-loss (Eqn.~\ref{eq:8}) also depends on the extent to which a reference architecture utilizes the formulation for better optimization. For SCAR~\cite{gao2019scar} and SFA-Net~\cite{zhu2019dual}, we hypothesize that this ability is relatively lower. Therefore, bin-loss is not always better for these networks (see Table~\ref{table:results}). Other networks (BL~\cite{ma2019bayesian}, DM-Count~\cite{wang2020DMCount}) utilize the loss better, leading to consistent improvement in MAE and standard deviation. However, SCAR~\cite{gao2019scar} is still better than no-binning in all cases except NWPU dataset. SFA-Net's performance with bin-loss included is better for the larger UCF, NWPU datasets. Also, inclusion of bin-loss results in consistent gains in terms of error standard deviation especially on the larger, heavily skewed datasets.

\subsubsection{Ablations on likelihood}

As mentioned in Sec.~\ref{sec:partitionprior}, we model the likelihood for each bin as a multinomial distribution. For comparative evaluation, we also consider two other candidate distributions for binning. The first candidate models the likelihood for the bin counts as a Poisson distribution: 

\begin{equation} \label{eqn:poibinlkhood}
\begin{split}
lik(B_k) & = lik(x_1,\dots , x_{m_k} ; \lambda_1, \dots, \lambda_{m_k}) \\
 & = \prod_{j=1}^{m_k} \frac{\lambda_j e^{-\lambda_j}}{x_j}
\end{split}
\end{equation}

where $\lambda_1, \dots, \lambda_{m_k}$ are the parameters of the Poission distributions associated with the bin elements. The other terms are used in the same context as Eqn.~\eqref{eqn:binlkhood} in Section~\ref{sec:partitionlkhood}. The second candidate distribution for binning is a variant of the multinomial, called stratified multinomial~\cite{sbb}. In this variant, the optimal Bayesian binning is applied not only to the count range, but also to the count frequency distribution. The comparative results can be seen in Table~\ref{table:likelihoodablation}. Though the pooled MAE with Poisson binning is slightly lower for random binning, the standard deviation is significantly larger than in the case of multinomial (as employed by us). The other results indicate the better overall stability arising from our simple yet effective choice for the likelihood distribution. 

\begin{table}[!t]
\caption{Ablations on the likelihood model for different choices of bin-level distribution. Though the pooled MAE with Poisson distribution is slightly lower for random binning, the standard deviation is significantly larger than our choice (multinomial).}
\label{table:likelihoodablation}
\centering
\resizebox{\linewidth}{!}
{
\centering

\begin{tabular}{c|c|c|c}
\toprule
     $Likelihood\downarrow$  Binning$\rightarrow$ & Bin Loss & Bin Loss (RR) & No-binning\\
\toprule
Poisson & $84.8^{\pm 441.2 }$ & $89.1^{\pm 533.1 }$ &  $77.8^{\pm 380.3 }$ \\
Stratified Multinomial & $90.0^{\pm 283.5 }$  & $90.6^{\pm 374.0 }$ & $80.7^{\pm 290.7 }$  \\
\textbf{Multinomial (ours)} &  $88.1^{\pm 236.7 } $ & $76.7^{\pm 205.0} $ & $77.8^{\pm 214.9 }$\\
\bottomrule
\end{tabular}
}
\end{table}

\section{Conclusion}

In this paper, we highlight biases at various stages of the typical crowd counting pipeline and propose novel modifications to address issues at each stage. We propose a novel Bayesian sample stratification approach to enable balanced minibatch sampling. Complementary to our sampling approach, we propose a novel loss function to encourage strata-aware optimization. We analyze the performance of crowd counting approaches across standard datasets and demonstrate that our proposed modifications reduce error standard deviation in a noticeable manner. Altogether, our contributions represent a nuanced, statistically balanced and fine-grained characterization of performance for crowd counting approaches.

The proposed bin-aware loss visibly reduces standard deviation of error. However, our work highlights the need for approaches in which error deviations are negligible compared to the mean error. We provide alternative measures of performance to compare performance of different approaches and measure their robustness across different datasets as well. Lastly, we use localization measures to get the complete picture of the performance.


We hope that our work motivates the community to join us in exploring these challenging aspects of the problem. Studying and addressing issues we have raised would enable statistically reliable crowd counting approaches in future.



\section*{Acknowledgment}

We would like to extend our gratitude to iHub-data for recognizing our work and offering a fellowship to one of the authors (Sravya).

\ifCLASSOPTIONcaptionsoff
  \newpage
\fi



\bibliographystyle{IEEEtran}
\bibliography{bibb}

\begin{thebibliography}{10}
\providecommand{\url}[1]{#1}
\csname url@samestyle\endcsname
\providecommand{\newblock}{\relax}
\providecommand{\bibinfo}[2]{#2}
\providecommand{\BIBentrySTDinterwordspacing}{\spaceskip=0pt\relax}
\providecommand{\BIBentryALTinterwordstretchfactor}{4}
\providecommand{\BIBentryALTinterwordspacing}{\spaceskip=\fontdimen2\font plus
\BIBentryALTinterwordstretchfactor\fontdimen3\font minus
  \fontdimen4\font\relax}
\providecommand{\BIBforeignlanguage}[2]{{%
\expandafter\ifx\csname l@#1\endcsname\relax
\typeout{** WARNING: IEEEtran.bst: No hyphenation pattern has been}%
\typeout{** loaded for the language `#1'. Using the pattern for}%
\typeout{** the default language instead.}%
\else
\language=\csname l@#1\endcsname
\fi
#2}}
\providecommand{\BIBdecl}{\relax}
\BIBdecl

\bibitem{mcnn}
Y.~Zhang, D.~Zhou, S.~Chen, S.~Gao, and Y.~Ma, ``Single-image crowd counting
  via multi-column convolutional neural network,'' in \emph{2016 IEEE
  Conference on Computer Vision and Pattern Recognition (CVPR)}, June 2016, pp.
  589--597.

\bibitem{idrees2018composition}
H.~Idrees, M.~Tayyab, K.~Athrey, D.~Zhang, S.~Al-Maadeed, N.~Rajpoot, and
  M.~Shah, ``Composition loss for counting, density map estimation and
  localization in dense crowds,'' in \emph{Computer Vision -- ECCV 2018},
  V.~Ferrari, M.~Hebert, C.~Sminchisescu, and Y.~Weiss, Eds.\hskip 1em plus
  0.5em minus 0.4em\relax Cham: Springer International Publishing, 2018, pp.
  544--559.

\bibitem{9009496}
\BIBentryALTinterwordspacing
V.~Sindagi, R.~Yasarla, and V.~Patel, ``Pushing the frontiers of unconstrained
  crowd counting: New dataset and benchmark method,'' in \emph{2019 IEEE/CVF
  International Conference on Computer Vision (ICCV)}.\hskip 1em plus 0.5em
  minus 0.4em\relax Los Alamitos, CA, USA: IEEE Computer Society, nov 2019, pp.
  1221--1231. [Online]. Available:
  \url{https://doi.ieeecomputersociety.org/10.1109/ICCV.2019.00131}
\BIBentrySTDinterwordspacing

\bibitem{gao2020nwpu}
Q.~Wang, J.~Gao, W.~Lin, and X.~Li, ``Nwpu-crowd: A large-scale benchmark for
  crowd counting and localization,'' \emph{IEEE Transactions on Pattern
  Analysis and Machine Intelligence}, 2020.

\bibitem{wang2020DMCount}
B.~Wang, H.~Liu, D.~Samaras, and M.~Hoai, ``Distribution matching for crowd
  counting,'' in \emph{Advances in Neural Information Processing Systems},
  2020.

\bibitem{Song_2021_ICCV}
Q.~Song, C.~Wang, Z.~Jiang, Y.~Wang, Y.~Tai, C.~Wang, J.~Li, F.~Huang, and
  Y.~Wu, ``Rethinking counting and localization in crowds: A purely point-based
  framework,'' in \emph{Proceedings of the IEEE/CVF International Conference on
  Computer Vision (ICCV)}, October 2021, pp. 3365--3374.

\bibitem{liang2021focal}
D.~Liang, W.~Xu, Y.~Zhu, and Y.~Zhou, ``Focal inverse distance transform maps
  for crowd localization and counting in dense crowd,'' \emph{arXiv preprint
  arXiv:2102.07925}, 2021.

\bibitem{Wan_2021_CVPR}
J.~Wan, Z.~Liu, and A.~B. Chan, ``A generalized loss function for crowd
  counting and localization,'' in \emph{Proceedings of the IEEE/CVF Conference
  on Computer Vision and Pattern Recognition (CVPR)}, June 2021, pp.
  1974--1983.

\bibitem{ma2019bayesian}
Z.~Ma, X.~Wei, X.~Hong, and Y.~Gong, ``Bayesian loss for crowd count estimation
  with point supervision,'' in \emph{Proceedings of the IEEE International
  Conference on Computer Vision}, 2019, pp. 6142--6151.

\bibitem{sasnet}
Q.~Song, C.~Wang, Y.~Wang, Y.~Tai, C.~Wang, J.~Li, J.~Wu, and J.~Ma, ``To
  choose or to fuse? scale selection for crowd counting,'' \emph{The
  Thirty-Fifth AAAI Conference on Artificial Intelligence (AAAI-21)}, 2021.

\bibitem{xhp2019SDCNet}
H.~Xiong, H.~Lu, C.~Liu, L.~Liang, Z.~Cao, and C.~Shen, ``From open set to
  closed set: Counting objects by spatial divide-and-conquer,'' in
  \emph{Proceedings of the IEEE/CVF International Conference on Computer Vision
  (ICCV)}, 2019, pp. 8362--8371.

\bibitem{gao2019scar}
\BIBentryALTinterwordspacing
J.~Gao, Q.~Wang, and Y.~Yuan, ``Scar: Spatial-/channel-wise attention
  regression networks for crowd counting,'' \emph{Neurocomputing}, vol. 363,
  pp. 1--8, 2019. [Online]. Available:
  \url{https://www.sciencedirect.com/science/article/pii/S0925231219311373}
\BIBentrySTDinterwordspacing

\bibitem{zhu2019dual}
\BIBentryALTinterwordspacing
L.~Zhu, Z.~Zhao, C.~Lu, Y.~Lin, Y.~Peng, and T.~Yao, ``Dual path multi-scale
  fusion networks with attention for crowd counting,'' \emph{CoRR}, vol.
  abs/1902.01115, 2019. [Online]. Available:
  \url{http://arxiv.org/abs/1902.01115}
\BIBentrySTDinterwordspacing

\bibitem{game}
R.~Guerrero-Gómez-Olmedo, B.~Torre-Jiménez, R.~López-Sastre,
  S.~Maldonado-Bascón, and D.~Oñoro, ``Extremely overlapping vehicle
  counting,'' 06 2015.

\bibitem{Scargle2013}
\BIBentryALTinterwordspacing
J.~D. Scargle, J.~P. Norris, B.~Jackson, and J.~Chiang, ``Studies in
  astronomical time series analysis. vi. bayesian block representations,''
  \emph{The Astrophysical Journal}, vol. 764, no.~2, p. 167, Feb 2013.
  [Online]. Available: \url{http://dx.doi.org/10.1088/0004-637X/764/2/167}
\BIBentrySTDinterwordspacing

\bibitem{daniel}
D.~Jurafsky and J.~H. Martin, \emph{Speech and Language Processing: An
  Introduction to Natural Language Processing, Computational Linguistics, and
  Speech Recognition}, 1st~ed.\hskip 1em plus 0.5em minus 0.4em\relax USA:
  Prentice Hall PTR, 2000.

\bibitem{sbb}
J.~Florjanczyk and T.~Sather, ``Stratified bayesian blocks,''
  \url{https://medium.com/@janplus/stratified-bayesian-blocks-2bd77c1e6cc7},
  Aug. 2015.

\end{thebibliography}
\end{document}